\documentclass[runningheads]{llncs}

\usepackage{eccv}

\usepackage{eccvabbrv}

\newcommand{\ours}[0]{RoMa~v2}

\usepackage{arydshln}
\usepackage{multirow}

\usepackage[all,british]{foreign}

\renewcommand\foreignabbrfont{\normalfont}

\DeclareRobustCommand{\vs}{\xperiodafter{{\foreignabbrfont{vs}}}}

\usepackage{mathtools}
\DeclareMathOperator{\softplus}{Softplus}
\DeclareMathOperator{\softmax}{Softmax}
\DeclarePairedDelimiterX{\abs}[1]{\lvert}{\rvert}{#1}
\DeclarePairedDelimiterX{\norm}[1]{\lVert}{\rVert}{#1}

\usepackage{bbold}

\makeatletter
\renewcommand\paragraph{\@startsection{paragraph}{4}{\z@}%
  {1mm \@plus1ex \@minus.2ex}%
  {-1em}%
  {\normalfont\normalsize\bfseries}}
\makeatother

\usepackage{pgfplots}
\pgfplotsset{width=7cm,compat=1.8}

\usepackage{graphicx}
\usepackage{booktabs}
\usepackage{subcaption}
\usepackage{wrapfig}
\usepackage{enumitem}
\usepackage{caption}

\usepackage[accsupp]{axessibility}  %

\usepackage{hyperref}

\usepackage{orcidlink}
\begin{document}

\title{RoMa v2: Harder Better Faster Denser Feature Matching}

\authorrunning{J. Edstedt et al.}
\institute{$^1$Linköping University \quad $^2$Chalmers University of Technology \quad $^3$University of Amsterdam \quad $^4$Centre for Mathematical Sciences, Lund University
}

\renewcommand{\thefootnote}{\fnsymbol{footnote}}

\author{
Johan Edstedt$^1$
\and
David Nordström$^2$
\and
Yushan Zhang$^1$
\and
Georg Bökman$^3$ \\
Jonathan Astermark$^4$ 
\and
Anders Heyden$^4$
\and
Viktor Larsson$^4$
\\
 Mårten Wadenbäck$^1$
\and
Michael Felsberg$^1$
\and
Fredrik Kahl$^2$
 \\
}

\maketitle

\begin{abstract}
Dense feature matching aims to estimate all correspondences between two images of a 3D scene and has recently been established as the gold standard due to its high accuracy and robustness.
However, existing dense matchers still fail or perform poorly for many hard real-world scenarios, and high-precision models are often slow, limiting their applicability.
In this paper, we attack these weaknesses on a wide front through a series of systematic improvements that together yield a significantly better model.
In particular, we construct a novel matching architecture and loss, which, combined with a curated diverse training distribution, enables our model to solve many complex matching tasks.
We further make training faster through a decoupled two-stage matching-then-refinement pipeline, and at the same time, significantly reduce refinement memory usage through a custom CUDA kernel. Finally, we leverage the recent DINOv3 foundation model along with multiple other insights to make the model more robust and unbiased.
In our extensive set of experiments, we show that the resulting novel matcher sets a new state-of-the-art, being significantly more accurate than its predecessors.\footnote{Code is available publicly at \href{https://github.com/Parskatt/RoMaV2}{https://github.com/Parskatt/RoMaV2}. }
\end{abstract}
    
\section{Introduction}
\label{sec:intro}

Dense matching is the task of matching every pixel of image $\mathbf{I}^A \in \mathbb{R}^{H^A \times W^A \times 3}$ with image $\mathbf{I}^B \in \mathbb{R}^{H^B \times W^B \times 3}$ in terms of a warp $\mathbf{W}^{A\mapsto B} \in \mathbb{R}^{H^A \times W^A \times 2}$ and a confidence mask $\mathbf{p}^{A\mapsto B} \in [0,1]^{H^A \times W^A \times 1}$. %
In dense \emph{feature} matching, the assumption is that the pixels in both images are observations of 3D points from the \emph{same scene}. %
In this case, for a perfect matcher, the confidence $\mathbf{p}^{A\mapsto B}$ is 1 for pixels corresponding to a 3D point in the scene that is observable from both views, i.e., that are co-visible, and $0$ for occluded pixels.

Feature matching is a fundamental task in Computer Vision, as many downstream tasks, e.g., visual localization~\cite{taira2018inloc,sattler2018benchmarking,panek2025guide} and 3D reconstruction~\cite{snavely2008modeling,opensfm,schoenberger2016sfm,lee2025dense,kotovenko2025edgs}, rely on precise and trustworthy correspondences in order to function robustly.
Traditionally, these methods relied on sparse matches established purely through descriptor similarity.
In the last couple of years, these detector-descriptor methods have been gradually replaced by learning-based matchers that consider pairs of images when establishing the correspondences, allowing the networks to not only consider visual similarity but also the spatial context.
\begin{figure*}
    \begin{subfigure}[t]{.33\linewidth}
        \centering
        \begin{minipage}[t][4.5cm][t]{\linewidth}
            \centering
            \raisebox{-\height}{\includegraphics[width=\linewidth]{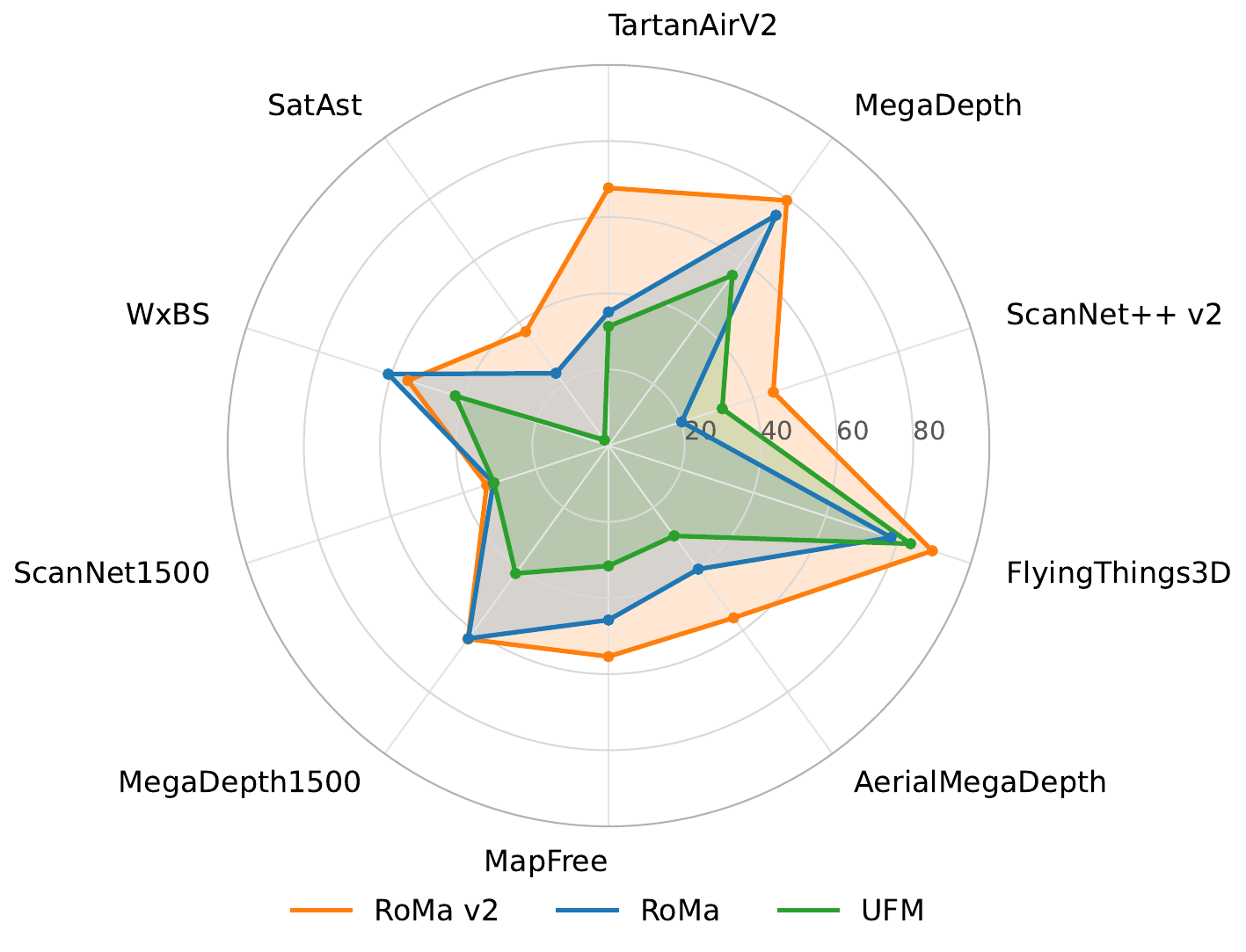}}
        \end{minipage}
        \caption{\textbf{Radar chart of performance on benchmarks.} Further details on these experiments can be found in~\Cref{sec:experiments}.}
        \label{fig:teaser}
    \end{subfigure}
    \hfill
    \begin{subfigure}[t]{.66\linewidth}
        \centering
        \begin{minipage}[t][4.5cm][t]{\linewidth}
            \centering
            \raisebox{-\height}{\includegraphics[width=0.245\linewidth]{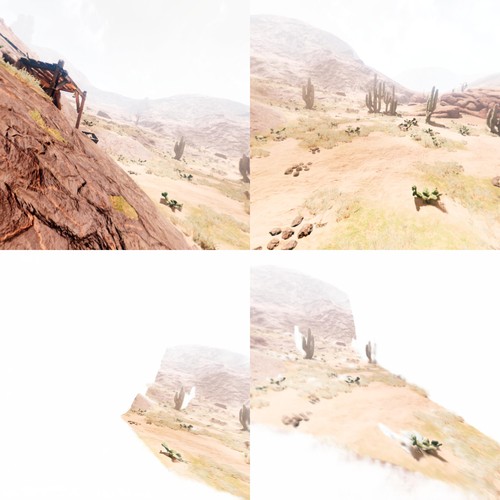}}\hfill%
            \raisebox{-\height}{\includegraphics[width=0.245\linewidth]{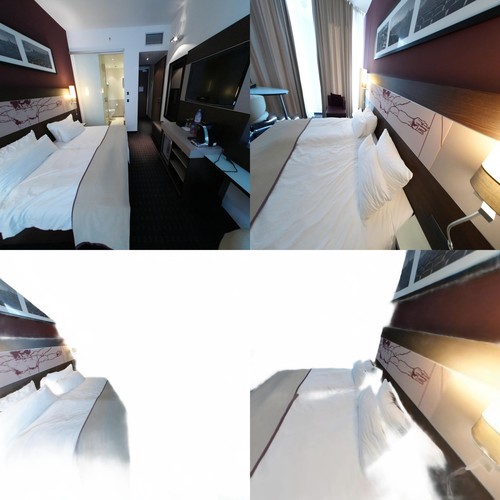}}\hfill%
            \raisebox{-\height}{\includegraphics[width=0.245\linewidth]{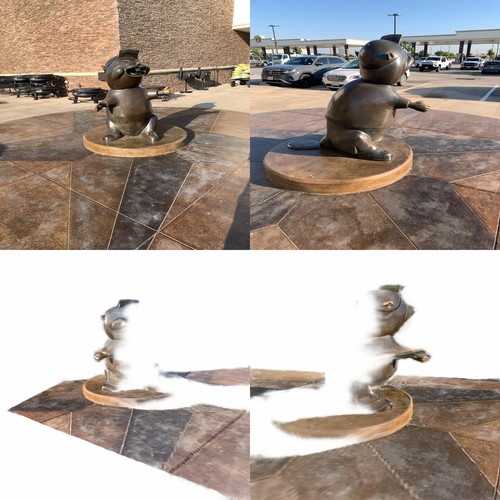}}\hfill%
            \raisebox{-\height}{\includegraphics[width=0.245\linewidth]{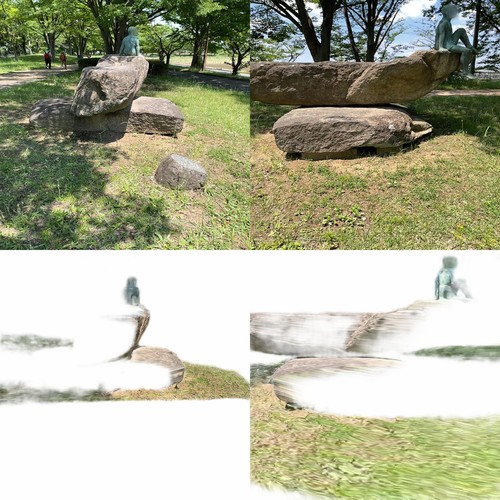}}\\[4pt]
            \raisebox{-\height}{\includegraphics[width=0.245\linewidth]{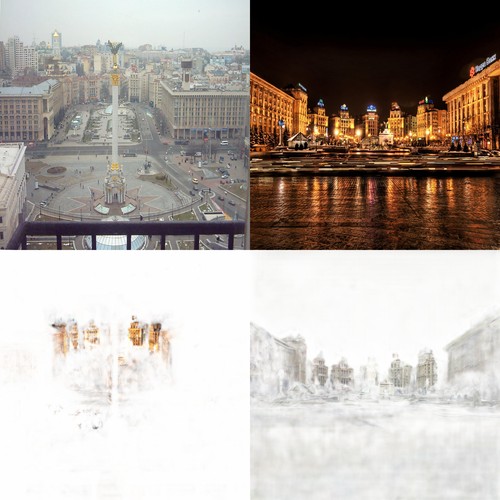}}\hfill%
            \raisebox{-\height}{\includegraphics[width=0.245\linewidth]{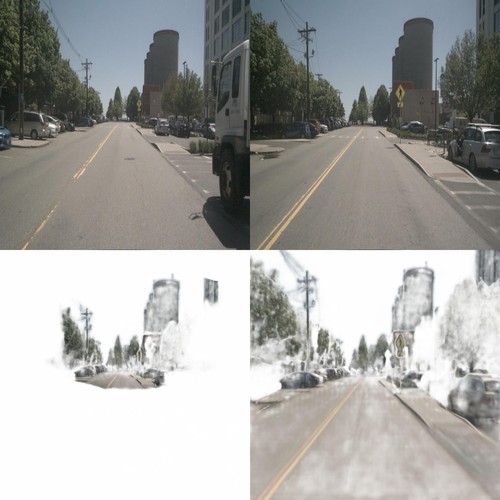}}\hfill%
            \raisebox{-\height}{\includegraphics[width=0.245\linewidth]{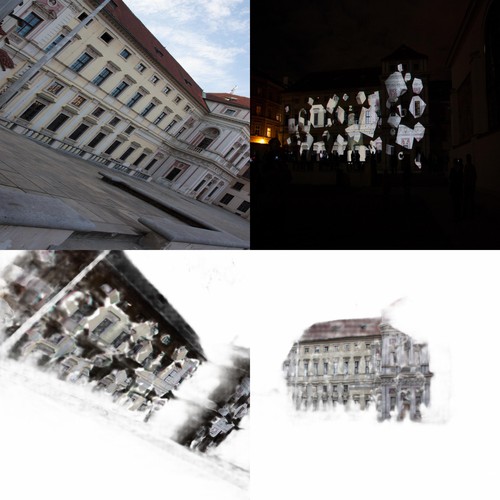}}\hfill%
            \raisebox{-\height}{\includegraphics[width=0.245\linewidth]{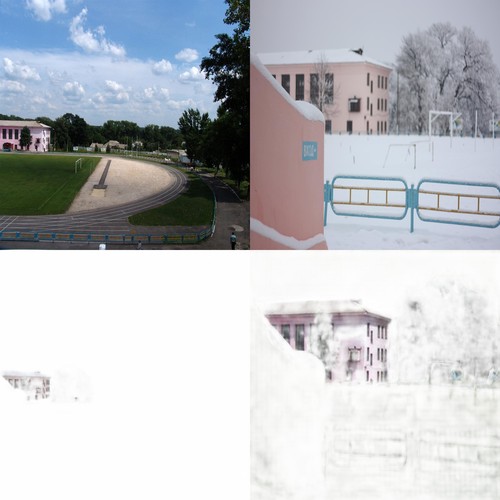}}
        \end{minipage}
        \caption{\textbf{Qualitative results.} Below each image pair we visualize the dense warp by coloring each pixel by the RGB value from its estimated corresponding location in the opposite image. Brighter values mean lower warp confidence as output by the model.}
        \label{fig:qualitative}
    \end{subfigure}
    \caption{\textbf{Quantitative and Qualitative results.} \ours~outperforms previous dense matchers on a wide range of pose estimation and dense matching tasks, as shown in \cref{fig:teaser}. We show a snapshot of results across different benchmarks in \cref{fig:qualitative}.} 
\end{figure*}

    This development in feature matching has been driven by the introduction of several challenging benchmarks, such as MegaDepth-1500~\cite{li2018megadepth,sun2021loftr}, ScanNet-1500~\cite{dai2017scannet,sarlin2020superglue}, WxBS~\cite{mishkin2015WXBS} and the recurring Image Matching Challenge at CVPR~\cite{image-matching-challenge-2022}.
    These benchmarks are currently dominated by detector-free methods, such as dense feature matchers~\cite{edstedt2023dkm} and feed-forward reconstruction models~\cite{wang2024dust3r}.
    
    One notable dense matcher is RoMa~\cite{edstedt2024roma}, which has proven robust to extreme photometric changes, including different modalities, due to using features from a frozen foundation model for the matching instead of learning features from scratch. 
    However, RoMa still struggles in many challenging scenarios.
    For example, the recent RUBIK benchmark~\cite{loiseau2025rubik}
    highlights its weakness under extreme viewpoint changes.
    Additionally, RoMa has a significant runtime and memory footprint, limiting its applicability for large-scale tasks or resource-constrained settings.
    Recently, UFM~\cite{zhang2025ufm} showed that dense matching can be made significantly faster than in RoMa.
    However, UFM requires finetuning of the pretrained feature extracting backbone, which leads to worse performance on datasets with extreme appearance changes such as WxBS.
    Furthermore, UFM performs worse than RoMa on benchmarks that require subpixel precision, such as MegaDepth-1500~\cite{li2018megadepth, sun2021loftr}.
    
    Motivated by the different trade-offs in RoMa and UFM, in this paper we address the challenge of combining their respective strengths, \ie, developing a dense feature matcher that is both robust to extreme changes in viewpoint and appearance, applicable to a wide range of real-world scenarios, all while maintaining subpixel precision and a practical runtime and memory footprint.
    To this end, we introduce \ours, which builds on RoMa and features several improvements to increase robustness while simultaneously reducing the computational cost.
    \ours~achieves state-of-the-art results on a wide range of benchmarks, as seen in Figure~\ref{fig:teaser}. Qualitative examples from the benchmarks are visualized in Figure~\ref{fig:qualitative}.

\noindent\textbf{In summary, our main contributions are:}
\begin{enumerate}%
    \item A novel matching objective, combining warp and correlation-based losses, which enables multi-view context to be learned in the coarse matcher of RoMa, described in~\Cref{sec:matcher}.
    \item Faster and less memory intensive refiners than in RoMa, described in~\Cref{sec:refinement} and ablated in~\Cref{tab:runtime}.
    \item A custom mixture of wide- and small-baseline datasets in the training data, that helps balance robustness to extreme viewpoints while maintaining sub-pixel performance across a wide range of difficult matching tasks, detailed in~\Cref{sec:data}.
    \item Prediction of pixel-wise error covariance that can be used downstream in refinement of estimated geometry, as demonstrated in~\Cref{sec:cov-experiments}.
    \item We experimentally verify that these improvements significantly reduce the runtime compared to baseline RoMa, while matching or outperforming both RoMa and UFM on their respective strong-points across a wide range of benchmarks in~\Cref{sec:experiments}.
\end{enumerate}

\section{Related Work}
\label{sec:rw}

\subsubsection{Feature Matching.}
Traditionally, feature matching has been dominated by the sparse paradigm, where keypoints are first detected separately in each image, and then matched by a sparse feature matcher.
While early approaches relied on similarity between local descriptors, the recent trend is instead to rely on a learned matcher that jointly considers the keypoints and descriptors in each image.
Notable recent works in this area include SuperGlue~\cite{sarlin2020superglue},  LightGlue~\cite{lindenberger2023lightglue}
and LoMa~\cite{nordstrom2026loma}, which all use an attention-based approach for solving the optimal assignment.
Common among the sparse methods is the reliance on salient image regions, where repeatable keypoints can be detected.
In contrast, LoFTR~\cite{sun2021loftr} takes a detector-free approach by matching through attention on learned features,
resulting in a semi-dense matching where even non-salient points can be matched.
\begin{figure*}
    \centering
    \includegraphics[width=0.85\linewidth]{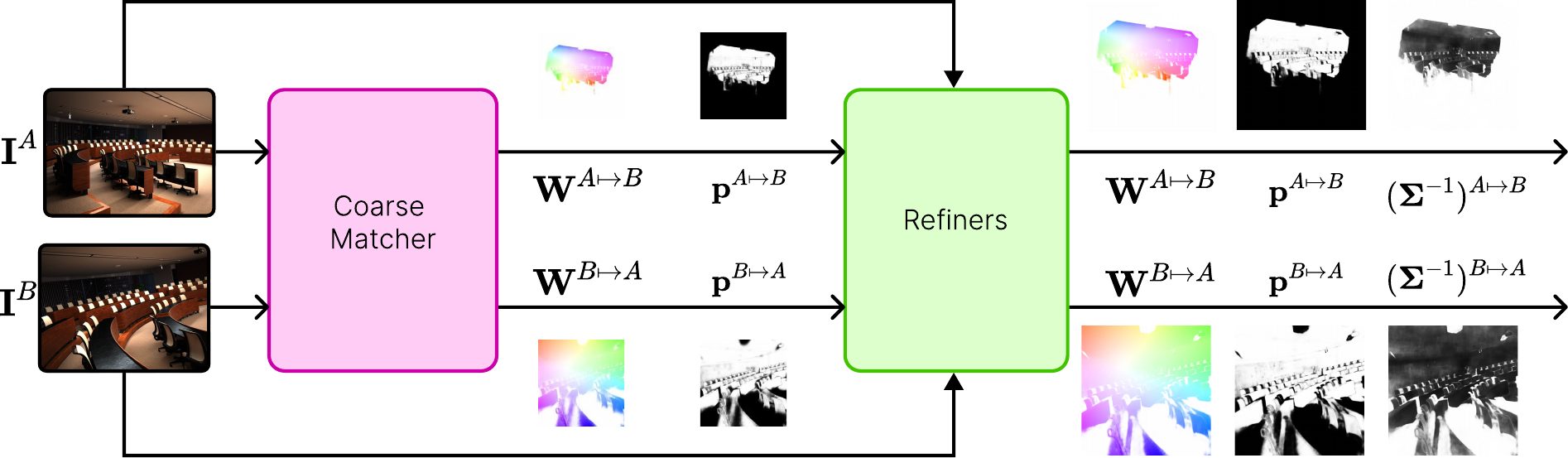}
    \caption{\textbf{Overview of \ours.} We estimate bidirectional dense image warps $\mathbf{W} = \{\mathbf{W}^{A\mapsto B}\in \mathbb{R}^{H\times W\times 2}, \mathbf{W}^{B\mapsto A}\in \mathbb{R}^{H\times W\times 2}\}$
    and warp confidences $\mathbf{p} = \{\mathbf{p}^{A\mapsto B}\in \mathbb{R}^{H\times W\times 1}, \mathbf{p}^{B\mapsto A}\in \mathbb{R}^{H\times W\times 1}\}$ between two input images using a two-stage pipeline consisting of a matching and refinement stage.
    Unlike recent SotA dense matchers, we additionally predict a precision matrix $\mathbf{\Sigma}^{-1} = \{(\mathbf{\Sigma}^{-1})^{A\mapsto B}\in \mathbb{R}^{H\times W\times 2 \times 2}, (\mathbf{\Sigma}^{-1})^{B\mapsto A}\in \mathbb{R}^{H\times W\times 2\times 2}\}$.
    The \textbf{coarse matcher} is a Multi-view Transformer that takes in frozen DINOv3~\cite{siméoni2025dinov3} foundation model features from image $\mathbf{I}^{A}\in\mathbb{R}^{H\times W \times 3}$ and $\mathbf{I}^{B}\in\mathbb{R}^{H\times W \times 3}$. Its internals are further illustrated in~\Cref{fig:coarse-matcher}, and explained in detail in~\Cref{sec:matcher}. 
    The \textbf{refiners} are fine-grained UNet-like CNN models that, conditioned on the previous warp and confidence, produce displacements and delta confidences. 
    Additionally, they predict a full $2\times 2$ precision matrix per pixel, which is visualized as $\abs[\big]{\mathbf{\Sigma}^{-1}}^{-1/4}$. 
    The refiners are further illustrated in~\Cref{fig:refiners} and explained in more detail in~\Cref{sec:refinement}.} 
    \label{fig:romav2}
\end{figure*}
In DKM~\cite{edstedt2023dkm}, the matching is performed on a pyramid of feature maps, enabling pixel-dense matches.
In a follow-up work, RoMa~\cite{edstedt2024roma} further improves on this method by using a frozen foundation model to encode the coarse matching features, making it significantly more robust to extreme appearance changes.
Recently, UFM~\cite{zhang2025ufm} was introduced as a more lightweight dense matcher, where the training of wide-baseline dense matching is unified with the related task of optical flow.

\subsubsection{Feed-forward Reconstruction.}
Recovering 3D structure and camera parameters from images, or Structure-from-Motion (SfM)~\cite{hartley2003multiple}, has traditionally relied on sequential pipelines such as Bundler~\cite{snavely2008modeling} and COLMAP~\cite{schoenberger2016sfm}, where point correspondences obtained through image matching play a central role.
Recently, learning-based SfM methods have emerged, and many now incorporate matching within a feed-forward architecture. DUSt3R~\cite{wang2024dust3r} and MASt3R~\cite{leroy2024grounding} directly regress point maps from image pairs; and VGGT~\cite{wang2025vggt} and MapAnything~\cite{keetha2025mapanything} extend this paradigm to longer sequences.
While these models can produce coarse correspondences, they struggle to yield accurate high-resolution dense matches. 
While our architecture is loosely inspired by these approaches, we retain an explicit dense matching formulation that provides subpixel-accurate correspondences.

\section{Method}
\label{sec:method}
In this section, we outline our proposed method. We begin by discussing our two-stage \textit{matching-then-refinement} architecture in Section~\ref{sec:arch} and proceed to discuss its two parts in \Cref{sec:matcher,sec:refinement}, respectively. In \Cref{sec:data}, we describe the training data used and in \Cref{sec:resolution} we explain the method used to make RoMa v2 more robust to changes in image resolution.

\subsection{Architecture}\label{sec:arch}
\begin{figure}
    \centering
    \includegraphics[width=\linewidth]{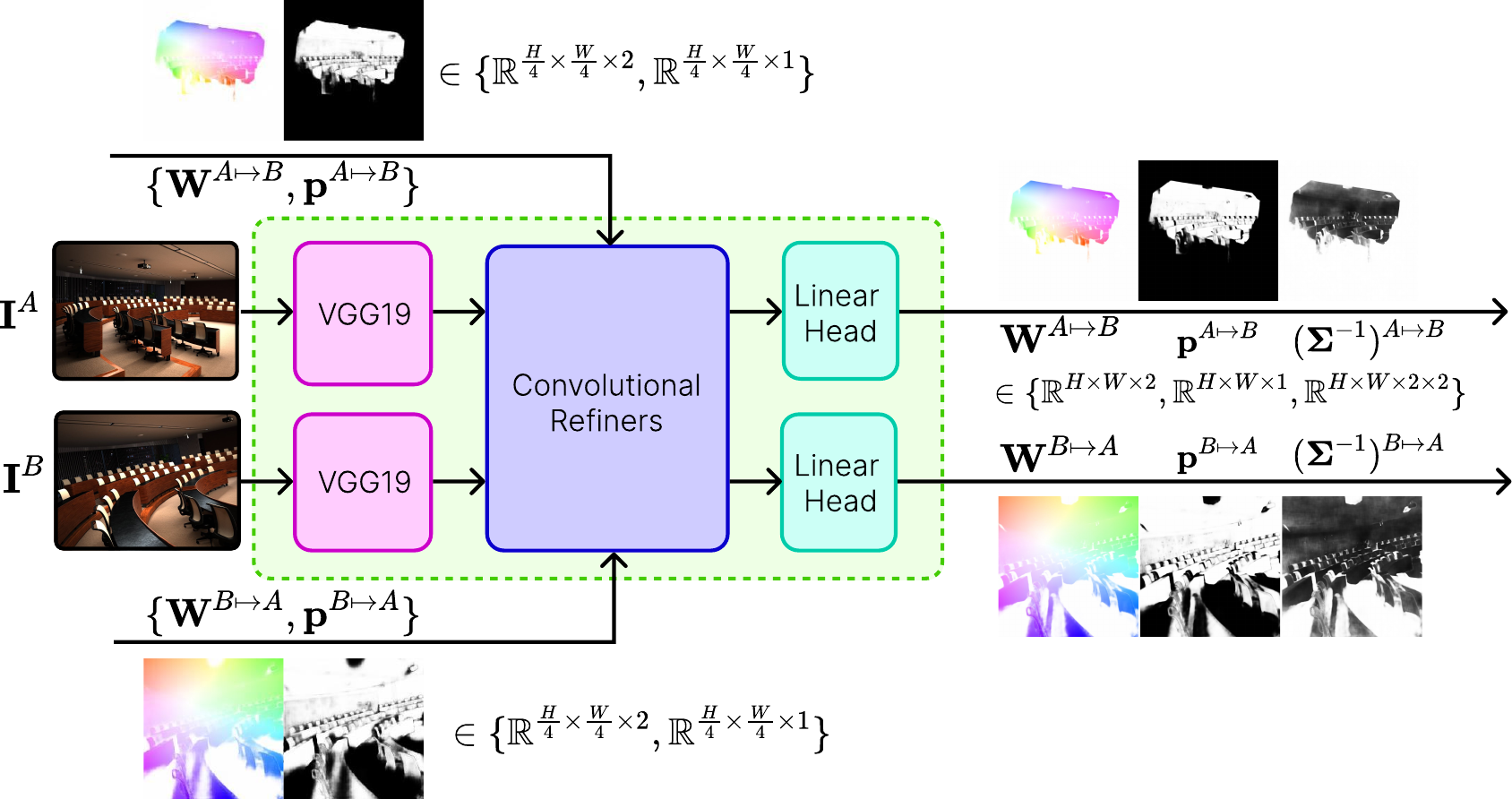}\vspace*{-0em}
    \caption{\textbf{Refiner internals.} The coarse matcher predicts at a resolution 4x smaller than the original image size. The refiners output at the original resolution.\vspace*{-1em}} %
    \label{fig:refiners}
\end{figure}

We take inspiration from previous works~\cite{edstedt2024roma,zhang2025ufm} and divide the dense matching task into a \emph{matching} step and a \emph{refinement} step. Intuitively, these two tasks entail first finding an approximate or \emph{coarse} match for each pixel, and, conditioned on this, refining the matching to sub-pixel accuracy.
An overview of the architecture is shown in~\Cref{fig:romav2}, and the respective components in~\Cref{fig:coarse-matcher} and~\Cref{fig:refiners}.

While some previous works, \eg \cite{edstedt2023dkm,edstedt2024roma}, decouple gradients between matchers and refiners but still train both jointly, we instead opt for a two-stage training paradigm inspired by UFM~\cite{zhang2025ufm}. 
This enables rapid experimentation. 
We next go into detail on the matcher in~\Cref{sec:matcher}, followed by the refinement in~\Cref{sec:refinement}.

\subsection{Matcher}
\label{sec:matcher}
\begin{figure}
    \centering
    \includegraphics[width=\linewidth]{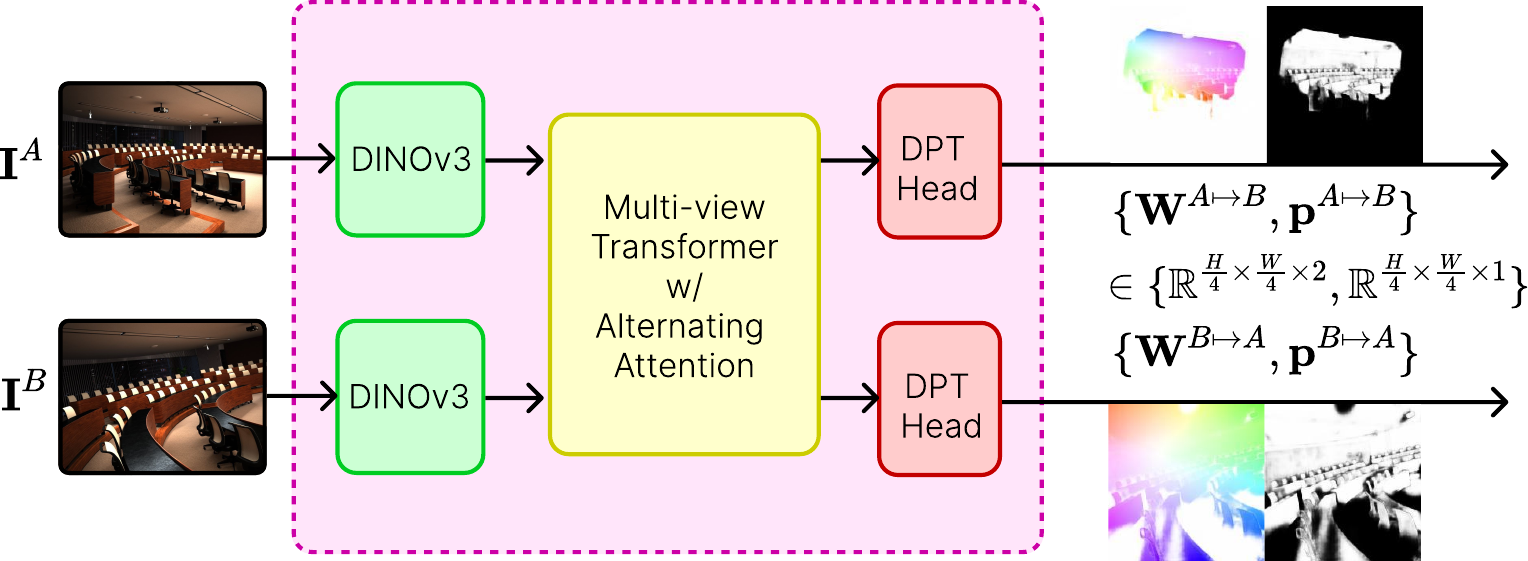}\vspace*{-.8em}
    \caption{\textbf{Coarse matcher.} We use a frozen DINOv3 feature extractor in the coarse matching stage. DINOv3 features from both input images are input to a Multi-view Transformer utilizing alternating Attention. Dense Prediction Transformer (DPT)~\cite{ranftl2021vision} heads output coarse warps $\mathbf{W}$ between the images and confidences $\mathbf{p}$ for 4x downsampled resolution.}
    \label{fig:coarse-matcher}
\end{figure}
An overview of the coarse matcher is shown in~\Cref{fig:coarse-matcher}. We upgrade the DINOv2~\cite{oquab2023dinov2} encoder used in RoMa to the newer DINOv3~\cite{siméoni2025dinov3}. Inspired by RoMa~\cite{edstedt2024roma}, we compare the encoders (frozen) by training a single linear layer on the features followed by a kernel nearest neighbor matcher. As is shown in Table~\ref{tab:linprobe}, we find that DINOv3 is more robust than its predecessor despite its slightly larger patch size (16 \vs 14). This is also supported by \Cref{tab:dinov3}. As in RoMa, but unlike UFM, we freeze the encoder weights.

While RoMa is robust and generalizes well, one of its main weaknesses is the lack of multi-view context in the matcher, which relies solely on Gaussian Process (GP)~\cite{rasmussen2003gaussian} regression combined with a single-view Transformer decoder to classify warp bins.
A naive approach would be to add a Multi-view Transformer to RoMa before the GP, however, we found that in practice the gradients through the GP were not sufficiently informative to yield improvements, and caused stability issues during training.

To remedy this, we replace the Gaussian Process with a simple single-headed Attention mechanism. 
Additionally, we add an auxiliary target, $\mathcal{L}_{\text{NLL}}$, to minimize the negative log-likelihood of the best matching patch in image B for each patch in image A. 
We first compute the similarity between all patches in image A and image B to form a similarity matrix $\mathcal{S} \in \mathbb{R}^{M \times N}$, where $M$ and $N$ are the number of patches in image A and B respectively. The loss $\mathcal{L}_{\text{NLL}}$ is computed by first applying $\softmax$ over the second dimension of $\mathcal{S}$ and then selecting the most similar pairs for all patches in A, \ie %
\begin{equation}
    \mathcal{L}_{\text{NLL}} = \sum_{m=1}^M - \log(
     \operatorname*{Softmax}(\mathcal{S}_m)_{n^*})%
\end{equation}
where $n^*$ is the index of the patch closest to the GT warp for patch $m$.
This approach can be seen as a dense directional version of, \eg LoFTR~\cite{sun2021loftr}.
However, we still use a regression head, which is trained to minimize the robust regression loss between its predicted warp and the ground truth warp.
The full coarse matching pipeline independently tokenizes the input images using DINOv3 ViT-L, and then applies a ViT-B Multi-view Transformer. Following  VGGT~\cite{wang2025vggt}, we alternate between frame-wise and global Attention. In contrast to VGGT~\cite{wang2025vggt}, we only use RoPE~\cite{rope} for the frame-wise Attention. The final token embeddings, $\softmax(\mathcal{S})x^B$, where $x^B$ are position embeddings as in RoMa, and DINOv3 features, are jointly processed by a Dense Prediction Transformer (DPT)~\cite{ranftl2021vision} head to predict the warp and confidence. Further details on the matcher architecture are given in the supplementary material.

\paragraph{Matching Loss:} We use the same overlap loss $\mathcal{L}_{\text{overlap}}$, and weighting factor ($\lambda=0.1$), as in RoMa. However, we replace the classification-by-regression term from the matching loss for the robust regression term $\mathcal{L}_{\text{warp}}$ used in the refinement loss. Finally, we add our proposed $\mathcal{L}_{\text{NLL}}$ and obtain:

\begin{equation}
    \mathcal{L}_{\text{matcher}} = \mathcal{L}_{\text{NLL}} + \mathcal{L}_{\text{warp}}(\mathbf{r}_{\theta_{\text{matcher}}}, \mathbf{p}_{\text{GT}}) + %
    \lambda
    \mathcal{L}_{\text{overlap}}(\mathbf{p}_{\theta_{\text{matcher}}}, \mathbf{p}_{\text{GT}}). \enspace
\end{equation}

In contrast to UFM, our matching objective incorporates the auxiliary target $\mathcal{L}_{\text{NLL}}$. %
We compare these architectures in Table~\ref{tab:ufm} by training on a subset of the data using the training setup outlined below and evaluating on holdout scenes from Hypersim~\cite{roberts2021hypersim}. We find that the \ours~setup works significantly better.

\begin{table}[ht] %
    \centering
    \begin{minipage}[t]{0.48\textwidth}
        \centering
\small
\caption{\textbf{Robustness of frozen features.} End-point-error (EPE) for a linear probe on MegaDepth. Robustness is the share of matches with error below 32px.}
\begin{tabular}{l rr}
    \toprule
    Method & EPE $\downarrow$ & Robustness \% $\uparrow$\\
    \midrule
      DINOv2 & 27.1 & 77.0 \\
      DINOv3 & \textbf{19.0} & \textbf{86.4}\\
      \bottomrule
\end{tabular}
\label{tab:linprobe}

    \end{minipage}
    \hfill
    \begin{minipage}[t]{0.48\textwidth}
        \centering
\caption{\textbf{Comparing \ours~and UFM matching architectures on Hypersim~\cite{roberts2021hypersim}}. Measured in PCK (higher is better).}
\begin{tabular}{l lll lll}
\toprule
 Method $\downarrow$\quad PCK$@$ $\rightarrow$
 &$1$px $\uparrow$&3px $\uparrow$& 5px $\uparrow$\\
 \midrule        
     UFM  &  11.2 & 48.3 & 67.4\\
     \ours & \bfseries 30.5 & \bfseries 76.7 & \bfseries 86.7 \\
\bottomrule
\end{tabular}

\label{tab:ufm}

    \end{minipage}
\end{table}
\paragraph{Coarse Matching Training:}
We start by training the coarse matcher for 300k steps of batch size 128, resulting in approximately 38M pairs seen throughout training. 
We use a learning rate of $4\cdot10^{-4}$.

\subsection{Refiners}
\label{sec:refinement}

After the matcher has finished training, we freeze it and run it in inference mode, producing a coarse warp for the refiners to refine.
We train the refiners for 300k steps of batch size 64, resulting in a total of approximately 19M pairs.
Like the matcher, we use a learning rate of $4\cdot10^{-4}$.

\paragraph{Architecture:}
As the matcher predicts at stride 4, in contrast to RoMa's stride 14, we only need to refine at strides $\leq 4$.
We thus construct three refiners at strides $\{4,2,1\}$, respectively.
These follow a similar architecture as in RoMa~\cite{edstedt2024roma} with some efficiency improvements.
First, we find that the local correlation implementation used in RoMa uses a large amount of memory, especially at high resolution.
To remedy this we write a custom CUDA kernel as a PyTorch extension, which significantly reduces the memory consumption (\cf~\Cref{tab:runtime}).
We further change all channel dimensions to be powers of two, which further boosts performance.
Further details about the %
refiners are given in the appendix.

\paragraph{Predictive Covariance:}
In addition to predicted overlap, it is often useful to have access to a numerical estimate of the expected error.
While predictive uncertainty has been previously studied~\cite{ilg2018uncertainty,schroppel2022benchmark,zhou2018deeptam,yin2019hierarchical,truong2023pdc,shenoi2026raco}, state-of-the-art matchers such as RoMa~\cite{edstedt2024roma} or UFM~\cite{zhang2025ufm} do not provide any such estimate.
To remedy this, we predict a pixel-wise Gaussian uncertainty of the 2D residuals,
\begin{equation}
\mathbf{r}_{\theta} := \mathbf{W}_{\theta}^{A\mapsto B} - \mathbf{W}^{A\mapsto B}_{\text{GT}} \in \mathbb{R}^{H\times W\times 2},
\end{equation}
through a $2\times 2$ precision matrix $\mathbf{P}_{\theta}\in \mathbb{R}^{H\times W\times 2\times 2}$
where $\mathbf{\Sigma}^{-1}_{\theta}(h,w) \succ 0$,
\ie, $\mathbf{\Sigma}^{-1}_{\theta}(h,w)$ is positive definite. We ensure this by constraining the network to predict the three elements $z_{11}, z_{21}, z_{22}$ and mapping these to Cholesky factors as $l_{11} = \softplus(z_{11})+10^{-6}$, $l_{21} = z_{21}$, $l_{22} = \softplus(z_{22})+10^{-6}$, where $\softplus(\cdot) = \ln(1+\exp(\cdot))$.
The lower triangular matrix is composed from the factors as
$
    L = \begin{pmatrix}
        l_{11} & 0 \\
        l_{21} & l_{22}
    \end{pmatrix}
$
and then the covariance matrix is formed from this as $\Sigma^{-1} = LL^{\top}$.
To learn $z_{11}, z_{21}, z_{22}$ we directly train the model to minimize the negative log-likelihood
\begin{equation}
    \mathcal{L}_{\text{prec}}(\mathbf{r}) = -\log\mathcal{N}(\mathbf{r}|0,\Sigma) = \frac{1}{2}\mathbf{r}^{\top}\Sigma^{-1}\mathbf{r} - \frac{1}{2}\log\det(\Sigma^{-1}) + \log(2\pi)
\end{equation}
To ensure stability, we only train the model to predict this covariance for covisible regions where $\norm{\mathbf{r}} < 8$ pixels.
We detach the residuals $\mathbf{r}_{\theta}$ before the loss.

We predict the precision in a hierarchical fashion from stride 4 up to stride 1, and use the fact that information is additive in the precision parameterization to predict our final precision matrix as
$
    \mathbf{\Sigma}^{-1}_{\theta_i} = \sum_{j \geq i}\Delta\mathbf{\Sigma}^{-1}_{\theta_j} .
$
We find empirically that our covariance improves performance in downstream tasks (\cf~\Cref{tab:hypersim-cov}), and that it qualitatively behaves as one would expect in~\Cref{fig:covariance_qualitative}.
\paragraph{EMA to remedy bias:}
\begin{figure}
    \centering
    \begin{subfigure}[t]{0.49\linewidth}
        \includegraphics[width=\linewidth]{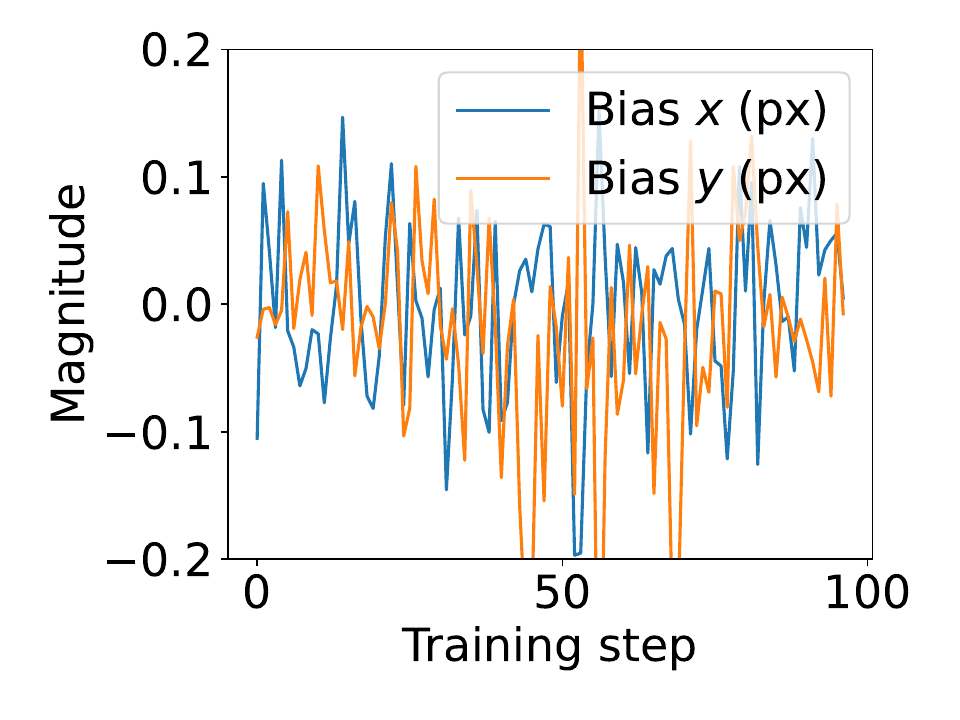}
        \caption{Before EMA}
        \label{fig:before-ema}
\end{subfigure}%
    \hfill 
    \begin{subfigure}[t]{0.49\linewidth}
\includegraphics[width=\linewidth]{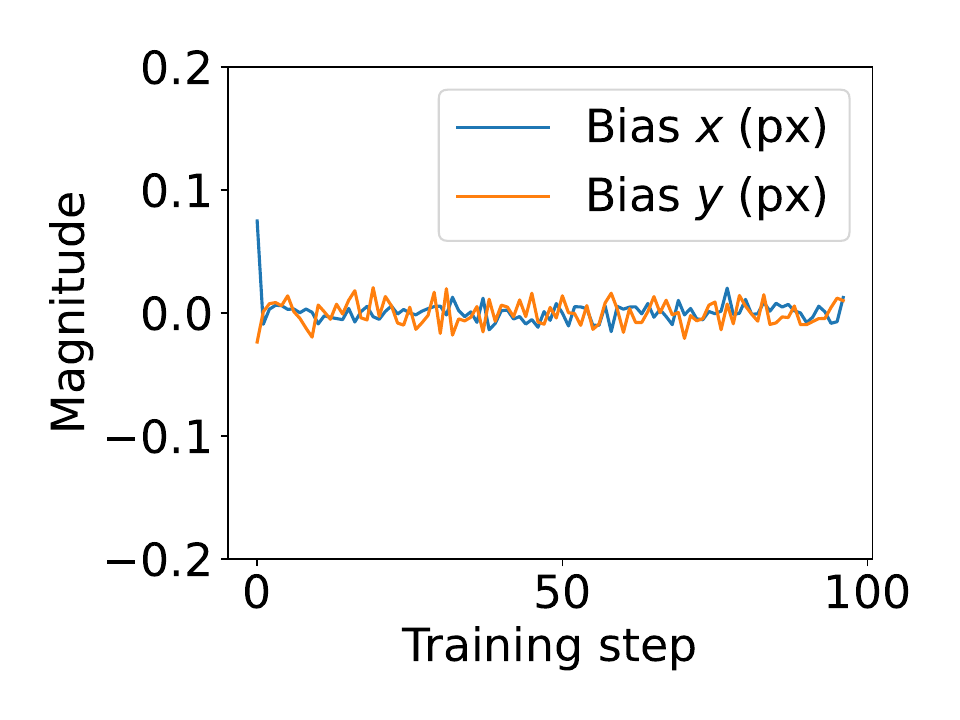}
        \caption{After EMA}
        \label{fig:after-ema}
\end{subfigure}
\caption{\textbf{Subpixel bias of refinement.} We observe that models exhibit subpixel fluctuations in their predictions throughout training, leading to bias. We propose a simple remedy through storing an exponential moving average (EMA).}
    \label{fig:bias}
\end{figure}
During training, we empirically observed that predictions tend to have a small, but noticeable, sub-pixel bias (typically around $\pm 0.1$ pixels in resolution $640\times 640$). At first, this seemed like a data issue, but after plotting this bias over the course of training we found that it appears almost random, see \Cref{fig:before-ema}. As the bias is seemingly uncorrelated over the course of training, a simple way to fix it is to use an Exponential Moving Average (EMA)\footnote{See~\cite{izmailov2018averaging} for discussion regarding different variants.}. We found a decay factor of $\alpha=0.999$ to work well empirically. After applying this remedy, we find that the bias in both orientations is substantially diminished, see \Cref{fig:after-ema}. We also find improved accuracy, see \Cref{tab:ema}.

\paragraph{Refinement Loss:}
We train the refiners using a combination of three losses.
Following RoMa we use a generalized Charbonnier loss~\cite{barron2019general} which for each refiner reads $\mathcal{L}_{\text{warp}} = (ic)^{\alpha}\left(\frac{\norm{\mathbf{r}}^2}{(ic)^2} + 1^2\right)^{\alpha/2}$, where we follow RoMa and set $\alpha = 0.5$, $c = 10^{-3}$ and $i\in\{4,2,1\}$ is the stride.

For estimating the overlap we follow UFM and RoMa and use a pixel-wise binary cross-entropy loss as $\mathcal{L}_{\text{ov}}$, with the ground truth overlap $\mathbf{p}_{\text{GT}} \in \{0,1\}$ being derived from either consistent depth (for MVS style datasets) or from warp cycle consistency (for flow datasets).
Further details on computing ground truth warps and overlaps are given in the suppl.~material. The total refinement loss is
\begin{equation}
        \mathcal{L}_{\text{refiners}} = \sum_{i\in \mathcal{S}} \mathcal{L}_{\text{warp}}(\mathbf{r}_{\theta_i}, \mathbf{p}_{\text{GT}}) + \lambda_{\text{ov}}\mathcal{L}_{\text{ov}}(\mathbf{p}_{\theta_i}, \mathbf{p}_{\text{GT}})+ \lambda_{\text{prec}}\mathcal{L}_{\text{prec}}(\mathbf{\Sigma}^{-1}_{\theta_i}, \text{detach}(\mathbf{r}_{\theta_i})),
\end{equation} where $\mathcal{S} = \{1,2,4\}$, $\lambda_{\text{ov}} = 10^{-2}$, $\lambda_{\text{prec}} = 10^{-3}$.

\subsection{Data}
\label{sec:data}
We train \ours~on a mix of wide and small baseline two-view datasets, a summary of which is presented in~\Cref{tab:dataset_mix}.
Our mix is inspired by UFM~\cite{zhang2025ufm}, and significantly more diverse than RoMa~\cite{edstedt2024roma}, which is only trained on MegaDepth.

\begin{figure}
\centering
\begin{subfigure}[t]{0.49\linewidth}
    \centering
    \includegraphics[width=.49\linewidth]{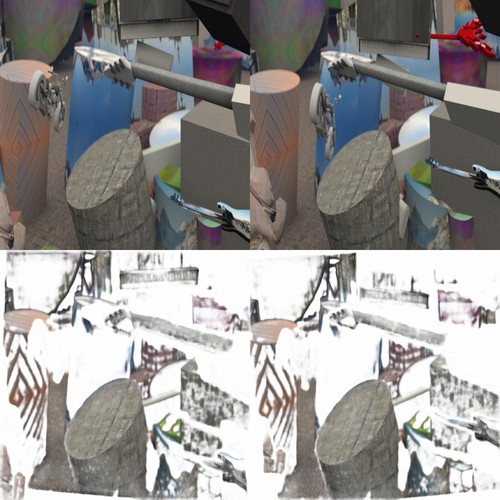}
    \includegraphics[width=.49\linewidth]{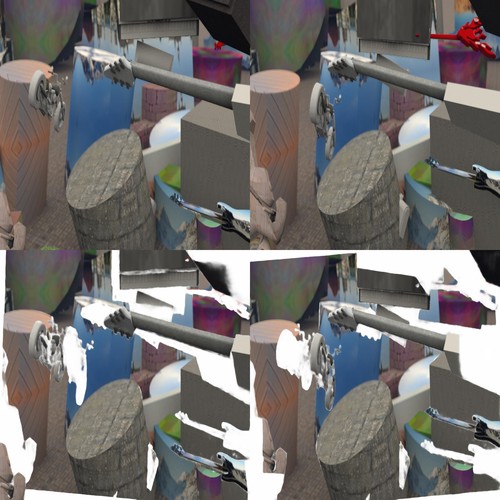}
\includegraphics[width=.24\linewidth]{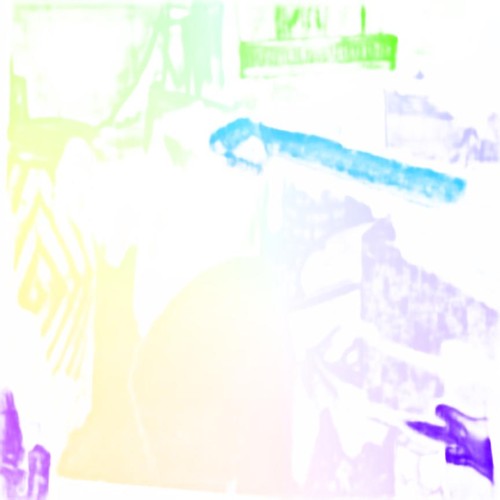}\includegraphics[width=.24\linewidth]{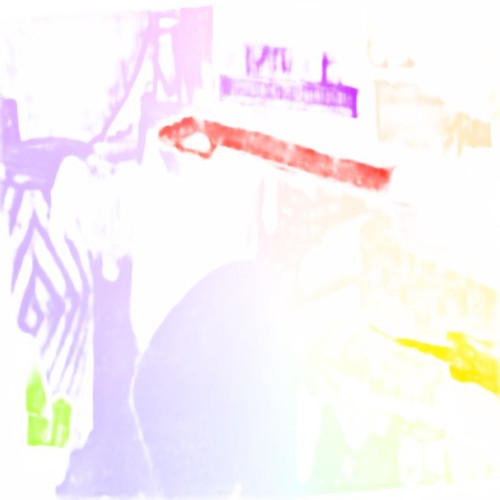}\includegraphics[width=.24\linewidth]{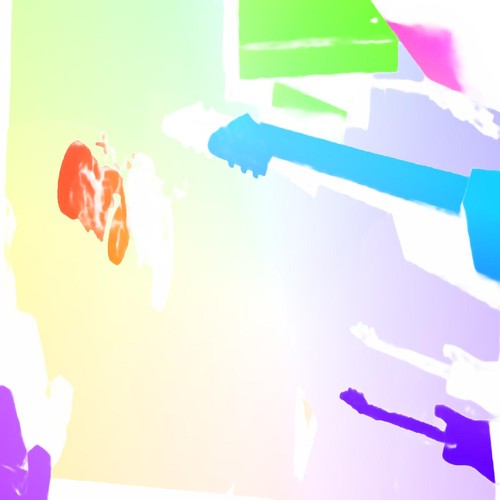} \includegraphics[width=.24\linewidth]{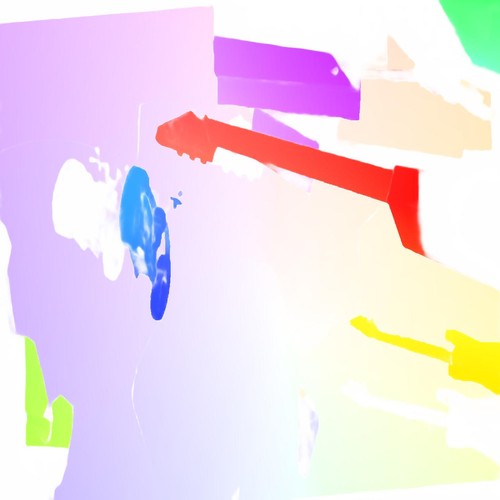}
    \caption{\textbf{Left:} RoMa warp for small baseline pair. Note the missing warp for the guitar in the bottom right. \textbf{Right:} \ours~warp. }
    \label{fig:flying-v2-v1}    
\end{subfigure}
\begin{subfigure}[t]{0.49\linewidth}
    \centering
\includegraphics[width=.49\linewidth]{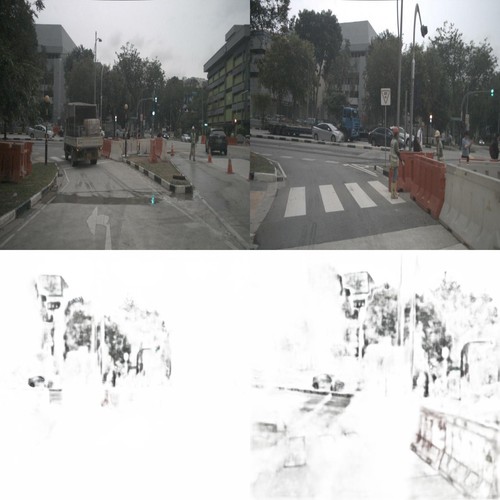} \includegraphics[width=.49\linewidth]{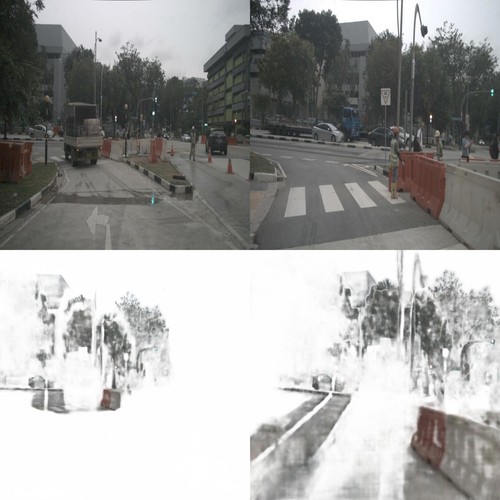} 
    \includegraphics[width=.24\linewidth]{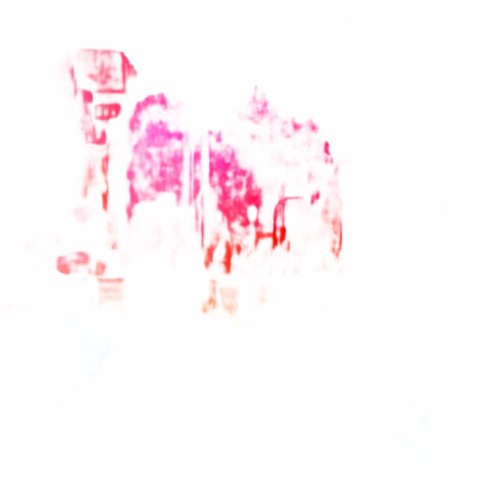}
    \includegraphics[width=.24\linewidth]{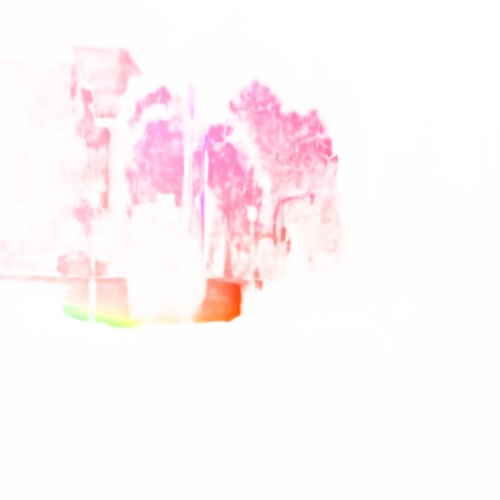}
\includegraphics[width=.24\linewidth]{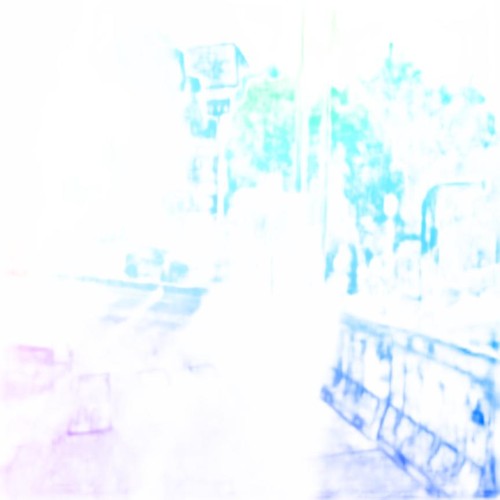}
\includegraphics[width=.23\linewidth]{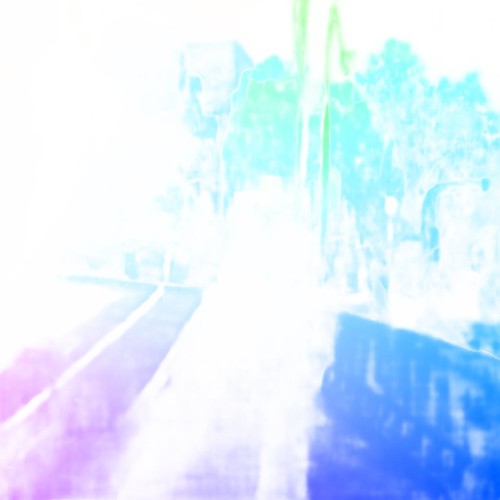}
    \caption{\textbf{Left:} RoMa warp for large-baseline pair with texture-poor road surfaces, with warp missing for almost the entire road. \textbf{Right:} \ours~warp. }
    \label{fig:nuscenes-v2-v1}
\end{subfigure}
\caption{\textbf{Qualitative Comparison.} \ours~is significantly better than RoMa at capturing small objects with dynamic motion \textbf{(a)}, and for texture-less scenes \textbf{(b)}}
\end{figure}

In particular, the inclusion of the Aerial datasets AerialMD~\cite{vuong2025aerialmegadepth} and BlendedMVS~\cite{yao2020blendedmvs}, enable our proposed model to be significantly more robust to large rotations and air-to-ground viewpoint changes. The inclusion of small-baseline datasets, like FlyingThings3D~\cite{mayer2016large}, makes \ours~significantly better at predicting fine-grained details.
We qualitatively compare \ours~to RoMa on fine-grained prediction on the FlyingThings3D dataset in~\Cref{fig:flying-v2-v1}.
We also find that our data mixture enables us to predict textureless surfaces significantly better than RoMa, particularly in Autonomous Driving (AD) scenarios, despite only training on the very small-scale dataset VKITTI2~\cite{gaidon2016virtual,cabon2020vkitti2}.
We demonstrate this for a randomly selected pair from the NuScenes dataset~\cite{caesar2020nuscenes} in~\Cref{fig:nuscenes-v2-v1}.

\begin{table}[ht] \centering
\scriptsize
\caption{\textbf{Dataset mixture} for {\ours }. The top part contains wide-baseline datasets, while the bottom part contains small-baseline datasets. The weight is proportional to the probability of sampling an image pair from the respective dataset. Further details about the datasets are provided in the supplementary.}
\label{tab:dataset_mix}
\setlength{\tabcolsep}{2pt}
\begin{tabular}{lccr}
\hline
\toprule
Datasets                          & Type / GT Source            & Weight & \textnumero Scenes \\ 
\midrule
MegaDepth~\cite{li2018megadepth}    & Outdoor / MVS        &1&  169 \\
AerialMD~\cite{vuong2025aerialmegadepth}    & Aerial / MVS        &1&  124 \\
BlendedMVS~\cite{yao2020blendedmvs}  & Aerial / Mesh      &1&  493 \\
Hypersim~\cite{roberts2021hypersim}& Indoor / Graphics          &1&  393 \\
TartanAir v2~\cite{wang2020tartanair}& Outdoor / Graphics          &1&  46 \\
Map-Free~\cite{arnold2022map}& Object-centric / MVS &1&  397 \\
ScanNet++ v2~\cite{yeshwanth2023scannet++} & Indoor / Mesh & 1 & 856 \\%\hdashline
\midrule
UnrealStereo4k~\cite{tosi2021unrealstereo4k} & Outdoor / Graphics & 0.01 & 8\\
Virtual KITTI 2~\cite{gaidon2016virtual,cabon2020vkitti2} & Outdoor / Graphics & 0.01 & 5\\
FlyingThings3D~\cite{mayer2016large} & Outdoor / Graphics & 0.5\phantom{0} & 2239\\

\midrule
\textbf{Total}                        &                     & & \textbf{5069} \\
\bottomrule
\end{tabular}
\vspace{-0.5em}
\normalsize
\end{table}

\subsection{Resolution}
\label{sec:resolution}
We find that some elementary key changes make matching and refinement robust to the choice of resolution.
\paragraph{Coarse Matching:}
Following DINOv3~\cite{siméoni2025dinov3} we use RoPE~\cite{rope} on a \emph{normalized} grid, rather than a pixel grid. 
This ensures that distances are always in distribution, even when changing resolution significantly.

Secondly, we find that the absolute position encodings used for the match embeddings need to be of low enough frequency to ensure that their interpolation is unproblematic~\cite{zhaoefficient}. %
Compared to RoMa, which initializes the scale to $\omega = 8$ and lets it be trained, we set it fixed to $\omega = 1$.
It is possible that the high frequency of the position embeddings is the cause of the issue that requires UFM to be run at a fixed resolution of $420\times560$ during inference.

\paragraph{Refinement:}
Ensuring resolution robustness for the refiners is non-trivial, as convolution is tied to the pixel grid. We find that the approach used in RoMa, whereby the input displacement is rescaled relative to a canonical resolution generalizes best.

\paragraph{Training:}
We train the coarse matcher on a mix of resolutions and aspect ratios, specifically: {\small \(\{512\times512,\,592\times448,\,624\times416,\,688\times384,\,448\times592,\,416\times624,\,384\times688\}\)}. The refiners are trained exclusively with size $640\times640$.

\section{Experiments}
\label{sec:experiments}
The qualitative improvements of \ours~as shown in~\Cref{fig:qualitative,fig:flying-v2-v1,fig:nuscenes-v2-v1} are confirmed by the quantitative results from extensive experiments, listed in this section.

\subsection{Relative Pose Estimation}
\label{sec:rel-pose}
We compare \ours~to state-of-the-art matchers and feed-forward 3D reconstruction methods on relative pose estimation. For sampling correspondences we follow RoMa and compute bidirectional warps from which we sample correspondences from a thresholded distribution where we set 
\begin{equation}
 \hat{\mathbf{p}}^{A\mapsto B} = \max(\mathbb{1}_{\mathbf{p}^{A\mapsto B} > 0.05},\mathbf{p}^{A\mapsto B}).
\end{equation}
We use a coarse resolution of $800\times800$ and a fine resolution of $1024\times 1024$.
Similarly, we sample a balanced subset of matches using their kernel density estimate approach.

We report results on MegaDepth-1500~\cite{li2018megadepth, sun2021loftr} and ScanNet-1500~\cite{dai2017scannet, sarlin2020superglue} in~\Cref{tab:relpose}. \ours\ consistently outperforms all prior matchers on both benchmarks. On MegaDepth, which demands accurate sub-pixel correspondences, \ours\ also surpasses all 3D reconstruction methods. 
On ScanNet, \ours\ achieves performance that is on par with VGGT and MASt3R.

\begin{table}[t]
\centering
\caption{\textbf{SotA comparison on MegaDepth-1500~\cite{li2018megadepth,sun2021loftr} and ScanNet-1500~\cite{dai2017scannet, sarlin2020superglue}}. The top part contains feed-forward 3D reconstruction models, while the bottom part contains feature matchers.}
\label{tab:relpose}
\small

\begin{tabular}{l r rrr r rrr}
\toprule
Method
&& \multicolumn{3}{c}{MegaDepth}
&& \multicolumn{3}{c}{ScanNet} \\
\cmidrule(lr){3-5} \cmidrule(lr){7-9}
AUC$@$ $\rightarrow$ 
&& $5^{\circ}$ & $10^{\circ}$ & $20^{\circ}$
&& $5^{\circ}$ & $10^{\circ}$ & $20^{\circ}$ \\
\midrule

MASt3R~\cite{leroy2024grounding}~\tiny{ECCV'24} && 42.4 & 61.5 & 76.9 && 33.6 & 56.8 & 74.1 \\

VGGT$^\dag$~\cite{wang2025vggt}~\tiny{CVPR'25} && 33.5 & 52.9 & 70.0 && 33.9 & 55.2 & 73.4\\

Reloc3r~\cite{dong2025reloc3r}~\tiny{CVPR'25} && \bfseries 49.6 & \bfseries 67.9 & \bfseries 81.2 && \bfseries 34.8 & \bfseries 58.4 & \bfseries 75.6 \\

\midrule           

LoFTR~\cite{sun2021loftr}~\tiny{CVPR'21}&& 52.8 & 69.2 & 81.2 && 22.1 & 40.8 & 57.6 \\

 DKM~\cite{edstedt2023dkm}~\tiny{CVPR'23}  &&  60.4 & 74.9 & 85.1 && 29.4 & 50.7 & 68.3 \\
 
 LightGlue~\cite{lindenberger2023lightglue}~\tiny{ICCV'23} && 51.0 & 68.1 & 80.7 && 17.8 & 34.0 & 52.0 \\ 
 
 RoMa~\cite{edstedt2024roma}~\tiny{CVPR'24} &&  62.6 & 76.7 & 86.3 && 31.8 & 53.4 & 70.9 \\

 UFM$^\dag$~\cite{zhang2025ufm}~\tiny{NeurIPS'25} && 41.5 & 57.9 & 72.4 && 31.3 & 54.1 & 72.0 \\

\ours && \bfseries 62.8 & \bfseries 77.0 & \bfseries 86.6 && \bfseries 33.6 & \bfseries 56.2 & \bfseries 73.8 \\
 
\bottomrule
\multicolumn{9}{l}{\scriptsize $^\dag$Our reproduced numbers.} \\
\end{tabular}
\end{table}

\subsection{Visual Localization}

We evaluate visual localization on Map-free~\cite{arnold2022map} and InLoc~\cite{taira2018inloc} in \Cref{tab:visloc}. For Map-free, we use the validation set and metric monocular depth from DA3~\cite{depthanything3}. We use the HLoc~\cite{sarlin2019coarse} pipeline to evaluate on InLoc~\cite{taira2018inloc}. \ours~significantly outperforms RoMa on both benchmarks. In fact, \ours~tops the public leaderboard for InLoc setting a new SotA. 

\begin{table}
\centering
\caption{\textbf{Visual Localization.} Comparison to RoMa on Map-free~\cite{arnold2022map} and InLoc~\cite{taira2018inloc}.}
\label{tab:visloc}
\small

\begin{tabular}{l r rr r cc}
\toprule
Method
&& \multicolumn{2}{c}{Map-free}
&& \multicolumn{2}{c}{InLoc} \\
\cmidrule(lr){3-4} \cmidrule(lr){6-7} 

&& Prec. & AUC
&& DUC1 & DUC2 \\
\midrule
RoMa && 59.7 & 84.4 && 60.6/79.3/89.9 & 66.4/83.2/87.8  \\
\ours && \bfseries 73.3 & \bfseries 89.6 && \bfseries 67.7/84.3/94.4 & \bfseries 73.3/86.3/90.8 \\
\bottomrule
\end{tabular}
\end{table}

\subsection{Dense Matching}

\begin{table*}
    \centering
    \caption{\textbf{Dense matching performance.}
    Images are resized to $640\times 640$.}
    \vspace{-.5em}
    \resizebox{\linewidth}{!}{%
    \begin{tabular}{l | rrrr | rrrr | rrrr}
    \toprule
        \multirow{2}{*}{\textbf{Method}}& \multicolumn{4}{ c }{\textbf{TA-WB}} & \multicolumn{4}{ c}{\textbf{MegaDepth}} & \multicolumn{4}{ c }{\textbf{ScanNet++ v2}} \\
          & EPE~$\downarrow$ & 1px~$\uparrow$  & 3px~$\uparrow$  & 5px~$\uparrow$  & EPE~$\downarrow$ & 1px~$\uparrow$  & 3px~$\uparrow$  & 5px~$\uparrow$  & EPE~$\downarrow$ & 1px~$\uparrow$  & 3px~$\uparrow$  & 5px~$\uparrow$   \\
         \midrule
     RoMa  & 60.61 & 35.1 & 52.6 & 56.2 & 2.34 & 74.8 & 93.7 & 96.4& 27.52 & 20.2 & 42.8 & 53.6 \\
     UFM  & 15.85 & 31.3 & 65.5 & 75.1 & 3.15 & 55.3 & 88.0 & 93.7& 6.93 & 31.4 & 67.7 & 80.0\\
     \ours  & \bfseries 13.82 & \bfseries 67.7 & \bfseries 81.8 & \bfseries 85.8 & \bfseries 1.47 & \bfseries 79.6 & \bfseries 94.7 & \bfseries 96.7& \bfseries 4.00 & \bfseries 45.5 & \bfseries 77.3 & \bfseries 86.6 \\
         \bottomrule
    \end{tabular}%
    }

    \label{tab:dense}
\end{table*}

We evaluate dense matching performance in \Cref{tab:dense} and \Cref{tab:dense2} on a wide array of datasets and compare with state-of-the-art dense matchers RoMa and UFM. For RoMa and \ours, we directly feed the $640\times 640$ images into the model, while for UFM we first resize the image to their suggested inference resolution $560\times 420$ and then bilinearly upsample the predictions back to $640\times 640$, as their precision degrades significantly for higher resolutions.

\ours~consistently outperforms previous methods across all 6 datasets. Notably, we simultaneously beat UFM on its own benchmark, TA-WB, and RoMa on MegaDepth, on which it is exclusively trained, while having $84\%$ lower EPE compared to RoMa on the challenging AerialMegaDepth benchmark.

\begin{table*}
    \centering
    \caption{\textbf{Further dense matching performance.}
    Images are resized to $640\times 640$.}
    \vspace{-.5em}
    \resizebox{\linewidth}{!}{%
    \begin{tabular}{l | rrrr | rrrr | rrrr}
    \toprule
        \multirow{2}{*}{\textbf{Method}}& \multicolumn{4}{ c }{\textbf{FlyingThings3D}} & \multicolumn{4}{ c}{\textbf{AerialMegaDepth}} & \multicolumn{4}{ c }{\textbf{MapFree}} \\
          & EPE~$\downarrow$ & 1px~$\uparrow$  & 3px~$\uparrow$  & 5px~$\uparrow$  & EPE~$\downarrow$ & 1px~$\uparrow$  & 3px~$\uparrow$  & 5px~$\uparrow$  & EPE~$\downarrow$ & 1px~$\uparrow$  & 3px~$\uparrow$  & 5px~$\uparrow$   \\
         \midrule
     RoMa  & 5.68 & 78.0 & 86.6 & 89.2 & 25.05 & 39.0 & 65.0 & 73.9& 8.55 & 45.8 & 72.3 & 80.9 \\
     UFM  & 1.33 & 83.4 & 93.9 & 96.1 & 17.44 & 29.3 & 61.6 & 73.8& 3.59 & 31.6 & 66.7 & 81.7\\
     \ours  & \bfseries 0.93 & \bfseries 89.4 & \bfseries 95.2 & \bfseries 96.8 & \bfseries 4.12 & \bfseries 55.9 & \bfseries 81.1 & \bfseries 87.9& \bfseries 2.03 & \bfseries 55.4 & \bfseries 84.9 & \bfseries 92.7 \\
         \bottomrule
    \end{tabular}%
    }
    \vspace{-0.5em}
    \label{tab:dense2}
\end{table*}

\subsection{Ablations and Runtime Comparisons}
In~\Cref{tab:runtime} we compare the runtime of~\ours~with RoMa and UFM.
As can be seen from the table, we improve the throughput significantly compared to RoMa, running $1.7\times$ faster.
Compared to UFM our model is slightly slower, however with a much smaller memory footprint.
In~\cref{tab:ema}, we ablate the refiner EMA using the same metrics as in~\cref{tab:relpose}; in~\cref{tab:dinov3} we ablate the backbone.
\begin{table}
    \centering
    \small
    \caption{\textbf{Runtime and memory.} Benchmarking 
    on $640\times 640$$^\dag$ images with a batch size of 8 on an H200. \ours~is $1.7\times$ faster than RoMa with similar memory footprint. $\mathcal{K}$ indicates the custom CUDA kernel for the local correlation operation.}
    \begin{tabular}{l rr}
        \toprule
        Method & Throughput (pairs/s) $\uparrow$ & Mem. (GB) $\downarrow$\\
        \midrule
            UFM & 43.0 & 16.2 \\
            RoMa & 18.5 & 4.7 \\
           \ours~(w/o $\mathcal{K}$) & 30.3 & 5.6 \\
          \ours~(w/ $\mathcal{K}$) & 30.9 & 4.8 \\
          \bottomrule
          \multicolumn{3}{l}{\footnotesize $^\dag$We use $644\times644$ for RoMa and UFM due to patch size 14.} \\

    \end{tabular}
    \vspace{-.5em}
    \label{tab:runtime}
\end{table}

\begin{table}[t]
    \centering
    \captionsetup{font=scriptsize}
    \small

    \begin{minipage}[t]{0.48\linewidth}
        \centering
        \caption{\textbf{EMA ablation.} Refiner EMA improves results specifically on benchmarks that require subpixel precision, such as Mega-1500.}
        \label{tab:ema}
        \begin{tabular}{l cc}
        \toprule
             & Mega-1500 $\uparrow$ & SN-1500 $\uparrow$ \\
             \midrule
             w/o EMA & (61.4, 75.8, 85.7) & (\textbf{33.6}, 56.1, 73.5)\\
             EMA & (\textbf{62.8}, \textbf{77.0}, \textbf{86.6}) & (\textbf{33.6}, \textbf{56.2}, \textbf{73.8})\\
        \bottomrule
        \end{tabular}
    \end{minipage}
    \hfill
    \begin{minipage}[t]{0.48\linewidth}
        \centering
        \caption{\textbf{Backbone ablation.} We compare using DINOv2 vs DINOv3 in the coarse matcher, skipping the refiners.}
        \label{tab:dinov3}
        \begin{tabular}{lcc}
        \toprule
        Backbone & WxBS & Hypersim \\
        \midrule
        DINOv2 & \bfseries 35.6 & 78.1 \\
        DINOv3 & 34.2 & \bfseries 79.2 \\
        \bottomrule
        \end{tabular}
    \end{minipage}

\end{table}

\subsection{Multi-Modal Matching on WxBS}
We evaluate the robustness of \ours~on the extremely challenging WxBS benchmark~\cite{mishkin2015WXBS}. 
This benchmark consists of hand-labeled correspondences between images taken with extreme changes in either viewpoint, illumination, modality, or all three, which measures the generalizability of the matcher to out-of-distribution downstream tasks.
Results are presented in~\Cref{tab:WxBS_SatAst}.
\begin{table}
\small
\centering
\caption{\textbf{SotA comparison on the WxBS~\cite{mishkin2015WXBS}, SatAst, and RUBIK~\cite{loiseau2025rubik} benchmarks.} mAA, AUC at 10px, and success ratio, respectively (higher is better). } 

\begin{tabular}{ l  r r r}
  \toprule
        Method & WxBS (mAA$@$10px) & SatAst (AUC$@$10px) & RUBIK (Success \%)\\
 \midrule
RoMa & \textbf{60.8} & 23.5 & 47.3 \\
UFM & 42.3 &  1.8 & 53.1 \\
\ours & 55.4 & \textbf{37.0} & \bfseries 57.3 \\
  \bottomrule
\end{tabular}
    \vspace{-1.5em}
\label{tab:WxBS_SatAst}
\end{table}

We observe that the performance of \ours~is significantly higher than UFM, but slightly lower than RoMa. 
Investigating the cause of this we found that \ours~and UFM both struggle with the IR-to-RGB multi-modal subset of WxBS.

\subsection{Astronaut to Satellite Image Matching}
We introduce a new benchmark, SatAst, for matching satellite images to images taken by astronauts from the International Space Station.
Prior work on this modality has focused on the retrieval task of searching a database of satellite images for the image content of a given astronaut image~\cite{stoken2023astronaut, berton2024earthloc, berton2024earthmatch, bökman2024steerers}.
We take 39 corresponding image pairs from EarthMatch~\cite{berton2024earthmatch} to create SatAst and hand-annotate 10 accurate correspondences for each pair.
Further, we include $90$ degree rotated copies of the satellite images, yielding a total of 156 image pairs. 
Given estimated correspondences from a model, we use RANSAC to obtain a homography and compute AUC@10px of the reprojection error of the annotated ground-truth correspondences using this homography.

SatAst is difficult, the most challenging aspects being i) satellite images are OOD for most matchers (including \ours), ii) large scale changes and iii) large in-plane rotations.
We compare \ours~against previous dense methods and present results in~\Cref{tab:WxBS_SatAst}.
More information about the benchmark is found in the supplementary material.

\subsection{Matching across Geometric Challenges on RUBIK}

We evaluate \ours~on RUBIK~\cite{loiseau2025rubik}, a subset of NuScenes~\cite{caesar2020nuscenes}, and report the success ratio (rotation error less than 5$^\circ$ and translation error less than 2m) in \cref{tab:WxBS_SatAst}. We set a new SotA of 57.3, beating the previous leader DUSt3R (54.8).

\subsection{Covariance Estimate}
\label{sec:cov-experiments}
\begin{figure*}
    \centering
    \begin{minipage}[t]{0.48\textwidth}
    \vspace{0pt}
        \centering
        \tiny
        \captionof{table}{\textbf{Impact of predictive covariance on Hypersim~\cite{roberts2021hypersim}}. Measured in AUC (higher is better). We use the predicted covariance to weight the residuals in the model refinement.}
        \begin{tabular}{l lll}
        \toprule
         Method $\downarrow$\quad\quad\quad AUC$@$ $\rightarrow$
         &$1^{\circ}$ $\uparrow$&$3^{\circ}$ $\uparrow$&$5^{\circ}$ $\uparrow$\\
         \midrule        
             \ours~(w/o $\Sigma^{-1}$) & 54.9 & 79.5 & 85.9 \\
             \ours~(w/ $\Sigma^{-1}$ Refine) & 75.8 & 89.0 & 92.6 \\
             \ours~(w/ $\Sigma^{-1}$ RANSAC + Refine) & \textbf{76.4} & \textbf{89.3} & \textbf{92.8} \\
        \bottomrule
        \end{tabular}
        \label{tab:hypersim-cov}
    \end{minipage}
    \hfill
    \begin{minipage}[t]{0.48\textwidth}
    \vspace{0pt}
        \centering
        \includegraphics[width=\linewidth]{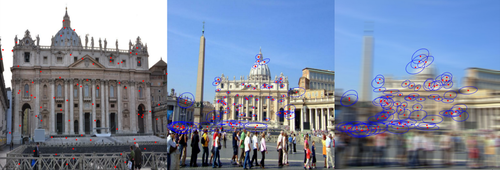}
        \captionof{figure}{\textbf{Qualitative example of covariances.} Predicted covariance for randomly sampled keypoints with simulated motion blur in the right image.}
        \label{fig:covariance_qualitative}
    \end{minipage}
\end{figure*}

While most robust pose estimation pipelines assume identically distributed residuals, our predictive covariance can be used to reweight residuals. %
To highlight the usefulness, we perform an experiment leveraging the covariance-weighted residuals. %
First, only as post-processing, refining the output of a classic point-based RANSAC.
Second, we compare with using it to reweight the residuals used for scoring inside RANSAC.
For the experiment, we consider image pairs sampled from the HyperSim~\cite{roberts2021hypersim} dataset.
In~\Cref{tab:hypersim-cov} we show that %
we gain significant improvements on 3D pose error metrics.
In particular, we improve by $\approx$ 20 points on AUC@1.
 Further experimental details %
 can be found in the supplementary material.
In~\Cref{fig:covariance_qualitative} we show a qualitative example of the predicted covariances. In the right image, we have applied a linear kernel to simulate motion blur, yielding larger covariances, especially in the blur direction.

\subsection{Importance of Subpixel Accuracy for MegaDepth-1500}
In~\Cref{sec:rel-pose} we claimed that subpixel accuracy is critical for high benchmark results on MegaDepth-1500.
To give evidence for this, we compute epipolar error histograms for RoMa v2 and UFM. 
Using these, we get a deterministic quantile mapping as
\begin{equation}
    \tilde{r}_{\text{RoMaV2 $\to$ UFM}} = Q_{\text{UFM}}(F_{\text{RoMa v2}}(r_{\text{RoMa v2}}))
\end{equation} where $Q$ is the residual quantile function of UFM and $F$ is the CDF of RoMa v2. We then simulate a less precise RoMa v2 by perturbing its predictions using the mapping. Results are shown in \Cref{tab:subpixel}.

\begin{table}[t]
\begin{center}
\small
\caption{Impact of subpixel accuracy.}\label{tab:subpixel}
\begin{tabular}{l rrr}
\toprule
Method 
& \multicolumn{3}{c}{\makebox[0pt]{MegaDepth-1500}}\\
\midrule
UFM & 41.5 & 57.9 & 72.4 \\
\midrule
RoMa v2 & {\bfseries 62.8} & {\bfseries 77.0} & {\bfseries 86.6} \\
RoMa v2 (Perturb.) & 46.3 & 63.7 & 77.4 \\
\bottomrule
\end{tabular}
\end{center}
\end{table}

\section{Limitations and Future Work}
\label{sec:limitations}
Compared to RoMa, our model is slightly less robust to extreme changes in modality, such as in WxBS\@. However, we are significantly more robust to these changes than UFM, as indicated by~\Cref{tab:WxBS_SatAst}.
Exploring the trade-offs between generalization and performance is an interesting direction for future work.
\section{Conclusion}
\label{sec:conclusion}
We have introduced \ours, a new dense feature matcher capable of matching \emph{harder} pairs, with \emph{better} (\ie, more precise) predictions, with \emph{faster} runtime than its predecessor, leading to \emph{denser} matches (\ie, \textit{more} correct matches).

\section*{Acknowledgements}
This work was supported by the Wallenberg Artificial
Intelligence, Autonomous Systems and Software Program
(WASP), funded by the Knut and Alice Wallenberg Foundation, and by the strategic research environment ELLIIT, funded by the Swedish government. 
The computational resources were provided by the
National Academic Infrastructure for Supercomputing in
Sweden (NAISS) at C3SE, partially funded by the Swedish Research
Council through grant agreement no.~2022-06725, and by
the Berzelius resource, provided by the Knut and Alice Wallenberg Foundation at the National Supercomputer Centre.

\bibliography{main}
\clearpage
\setcounter{page}{1}
\setcounter{section}{0}

\renewcommand{\thesection}{\Alph{section}}

\begin{center}
{\Large\bf \ours: Harder Better Faster Denser Feature Matching}\\[0.5em]
{\large Supplementary Material}
\end{center}

\section{Architectural Details}
Here we give further details on the exact dimensions of inputs and outputs of the different components of our model.
\paragraph{Matcher:}
The matcher takes in a list of features from the DINOv3 ViT-L backbone, in our implementation specifically layers 11 and 17, each have dimension $D_{\text{DINO}}=1024$, and which we denote as $\mathbf{f}_{\{11,17\}}^A, \mathbf{f}_{\{11,17\}}^B$.
These features are concatenated into a $2048$-dimensional feature, which is linearly projected into a $768$-dimensional subspace as 
\begin{align}
    \tilde{\mathbf{f}}^A &= P (\mathbf{f}_{11}^A \oplus \mathbf{f}_{17}^A)\in\mathbb{R}^{768}, P\in \mathbb{R}^{768\times 2048} \\
        \tilde{\mathbf{f}}^B &= P (\mathbf{f}_{11}^B \oplus \mathbf{f}_{17}^B)\in\mathbb{R}^{768}, P\in \mathbb{R}^{768\times 2048}
\end{align}
The projected features from image A and image B are then stacked and fed into an alternating Attention Multi-view Transformer of ViT-B architecture (we use a standard implementation with dim=768, depth=12, num\_heads=12, ffn\_ratio=4, and do not employ LayerScale, we however retain the $1024$ output dim through a linear map to conform to ViT-L) as
\begin{equation}
    (\textbf{z}^A, \textbf{z}^B) = m_{\theta}(\tilde{\mathbf{f}}^A,\tilde{\mathbf{f}}^B) \in (\mathbb{R}^{M\times 1024}, \mathbb{R}^{N\times 1024})
\end{equation}
This Transformer alternates between global Attention, processing both frames jointly without any positional encoding, and frame-wise Attention using normalized Axial RoPE (as in DINOv3).

The output of $m_{\theta}$ is used to construct the similarity matrix $\mathcal{S}\in \mathbb{R}^{M\times N}$ as
\begin{equation}
    \mathcal{S}_{mn} = \exp(1/\tau\text{ cossim}(\mathbf{z}^A_m, \mathbf{z}^B_n))
\end{equation}
where $\tau=1/10$ is the temperature following RoMa, and cossim denotes cosine similarity, i.e., $\text{cossim}(x,y) = \frac{x\cdot y}{\norm{x}\,\norm{y}}$ where $\cdot$ is the dot product.
Using this similarity matrix, we compute so-called ``match embeddings'' (following the nomenclature of RoMa) as 
\begin{equation}
    \chi^{A\mapsto B}_m = \mathcal{S}_m\chi^{B} \in \mathbb{R}^{1024},
\end{equation}
where $\chi_n^B = \cos(2\pi\omega Wx_n^B) \oplus \sin(2\pi\omega Wx_n^B)\in \mathbb{R}^{1024}$, $x_n^B\in\mathbb{R}^2$ is the pixel-coordinate of patch $n$ in image $B$, $\omega = 1$ (as discussed in the main text), and $W\in \mathbb{R}^{512\times2}$ is a non-learnable matrix with elements drawn from $\mathcal{N}(0,1)$.
We combine features into input for a DPT~\cite{ranftl2021vision} head as
\begin{equation}
    \{\mathbf{f}_{11}^B , \mathbf{f}_{11}^B, \mathbf{f}_{17}^B + \chi^{A\mapsto B}_m + \mathbf{z}^A,\mathbf{f}_{17}^B + \chi^{A\mapsto B}_m + \mathbf{z}^A\},
\end{equation}
where we set the finest resolution to a quarter of the original image size. 
We use a scratch dimension of $256$ and out dimensions of $[256,512,1024,1024]$ for strides $[4,8,16,32]$ respectively for the DPT head. The final prediction is made at stride $4$.

\paragraph{Refiners:}
We use a modified version of the refiners proposed in DKM and RoMa~\cite{edstedt2023dkm,edstedt2024roma}.
In particular, we retain only the refiners at stride $[4,2,1]$, due to matching at stride $4$.
This has the effect of making the refinement significantly cheaper, as we also only have to extract features from the VGG19 backbone until stride 4, compared to RoMa and DKM which require features and refinement from stride $16$.
We denote the fine features as
\begin{equation}
    \mathbf{\varphi}_4\in \mathbb{R}^{\frac{H}{4}\times\frac{W}{4}\times 192}, \mathbf{\varphi}_2 \in \mathbb{R}^{\frac{H}{2}\times\frac{W}{2}\times 48}, \mathbf{\varphi}_1 \in  \mathbb{R}^{H\times W\times 12},
\end{equation}
where the dimensions come from the raw VGG19 features (extracted right before the corresponding maxpool) projected with linear layers of sizes $\mathbb{R}^{192\times 256},\mathbb{R}^{48\times 128},\mathbb{R}^{12\times 64}$.
Refiners at stride $i$ take input of the form.
\begin{align}
    \mathbf{\varphi}_i^A &\oplus \mathbf{\varphi}_i^B(\mathbf{W}^{A\mapsto B}) \oplus g_i(\mathbf{W}^{A\mapsto B} - \mathbf{x}^A) \\&\oplus \text{ local\_corr}(\mathbf{\varphi}_i^A, \mathbf{\varphi}_i^B, \mathbf{W}^{A\mapsto B}, k_i)
\end{align}
where at each pixel $x^a$ local\_corr uses the previous warp to construct a $k_i\times k_i$ local correlation around $\mathbf{W}^{A\mapsto B}(x^A)$, and $g_i$ are linear maps.
We use $[k_4, k_2,k_1] = [7, 3,0 \text{ (no corr)}]$, $g_4 = \mathbb{R}^{79\times 2}, g_2 = \mathbb{R}^{23\times 2}, g_1 = \mathbb{R}^{8\times 2}$.
The concatenation of all these features sum up to $512, 128, 32$ respectively for strides $4,2,1$, which are intentionally powers of two, as this slightly increases inference speed. The internals are as in DKM and RoMa, that is, 8 layers each consisting of $5\times 5$ depthwise convolution, followed by BatchNorm, ReLU, and $1\times 1$ pointwise convolution.

\section{Further Details on Datasets}
\paragraph{MegaDepth and AerialMegaDepth:}
We follow the setup in RoMa~\cite{edstedt2024roma}, which is the following.
For each scene, directional overlaps are first computed using the number of shared 3D tracks divided by the number of observed tracks, giving a number between 0 and 1.
Up to 200000 pairs are selected from each scene by randomly sampling up to 100000 pairs with overlap $> 0.01$, and up to 100000 pairs with overlap $>0.35$.
Different from other datasets, sampling is not done uniformly over scenes. 
Rather, sampling is done over pairs, but pair sampling likelihood is down-weighted by the number of pairs in the scene to the power of $0.75$. 
Note that if the power had been $1$ this would be equivalent to uniform sampling.
\paragraph{MapFree}
We run COLMAP's MVS on all scenes with default settings, giving us per image depth maps.
Like MegaDepth we compute overlaps as the directional percentage of shared 3D tracks between images.
For training we sample pairs with overlap $>0.01$ uniformly over the scenes, using only seq0 per-scene.

\paragraph{ScanNet++ v2:}
We train on the \texttt{nvs\_sem\_train} split which consists of 856 indoor scenes from which we use the DSLR images.
For each scene we render image aligned depth maps from the scene mesh, which is derived from a Faro Focus Premium laser scanner.
We compute the overlap pairs of images as the geometric mean of their respective directional depth map overlaps in $512\times 512$ resolution.
We use a threshold of 0.2 for the minimum required overlap.
For each scene we compute 10000 pairs that fulfill the overlap threshold, and use them for training. This gives us $\approx8\cdot10^6$ total pairs.

\paragraph{TartanAir V2:}
We follow the setup in UFM and use their TA-WB pairs for training, where we like UFM leave out the OldScandinavia,
Sewerage, Supermarket, DesertGasStation, and PolarSciFi scenes for test.
For further details about the pair construction, see UFM~\cite{zhang2025ufm}.
\paragraph{BlendedMVS:}
We follow MapAnything and exclude the scenes:
\begin{itemize}[label=-,labelindent=1em,before=\footnotesize\ttfamily]
    \item 5692a4c2adafac1f14201821,
    \item 5864a935712e2761469111b4,
    \item 59f87d0bfa6280566fb38c9a,
    \item 58a44463156b87103d3ed45e,
    \item 5c2b3ed5e611832e8aed46bf,
    \item 5bf03590d4392319481971dc,
    \item 00000000000000000000001a,
    \item 00000000000000000000000c,
    \item 000000000000000000000000.
\end{itemize}
We train on all other scenes.
We use pairs with directional overlap (computed from the depth maps) larger than $0.05$.

\paragraph{Hypersim:}
We train on scenes with index $<50$, and validate on scenes with index $\ge 50$.
We sample pairs with a unidirectional overlap (based on depth maps) $\ge 0.2$.

\paragraph{FlyingThings3D:}
We use the official ``TRAIN'' ``TEST'' split and train on both the ``clean'' and ``final''-pass images, converting the provided optical flows into warps.

\paragraph{UnrealStereo4K:}
We train on all scenes and use the left/right images with their disparities, which we convert to warps through the disparity/depth inverse relation.

\paragraph{Virtual KITTI 2:}
We train on all scenes, using subsequent frames with the same condition and camera, and deriving the warp from the provided optical flow.
During training we randomly draw conditions (weather conditions and camera rig position), and using either left or right stereo camera.

\section{Further Details on Training Data}
\begin{figure*}
    \centering
\begin{subfigure}[t]{0.49\linewidth}
    \centering
    \includegraphics[width=0.235\linewidth]{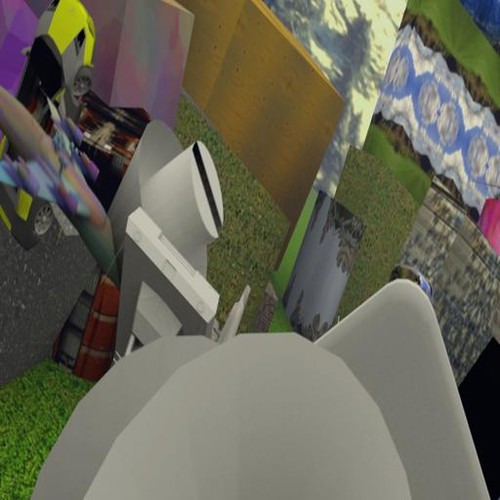}
    \includegraphics[width=0.235\linewidth]{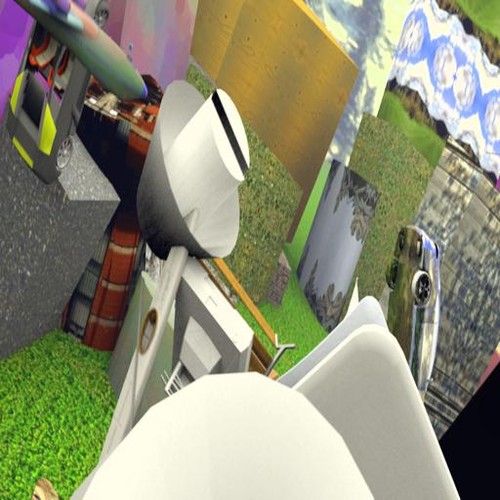}
    \includegraphics[width=0.235\linewidth]{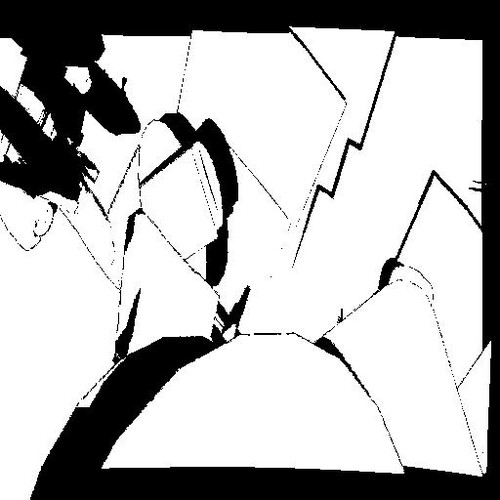}
    \includegraphics[width=0.235\linewidth]{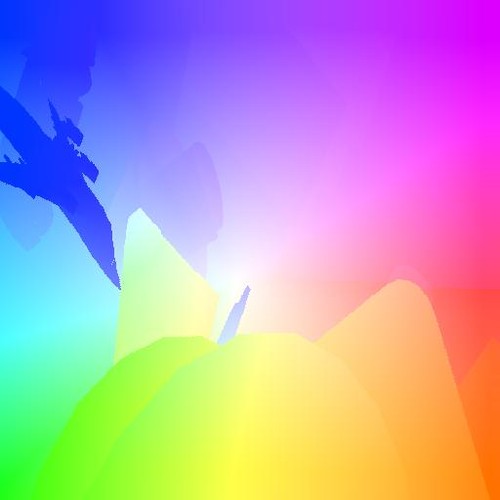}

    \includegraphics[width=0.235\linewidth]{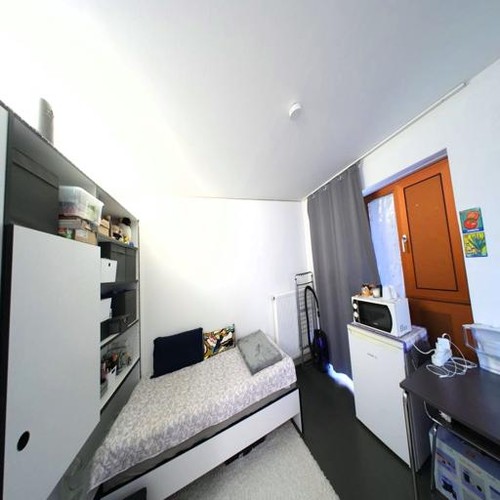}
    \includegraphics[width=0.235\linewidth]{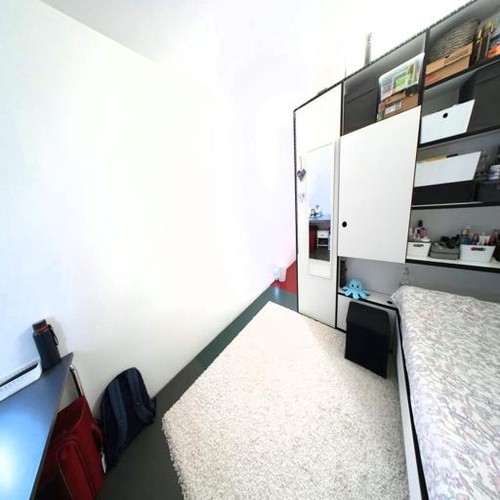}
    \includegraphics[width=0.235\linewidth]{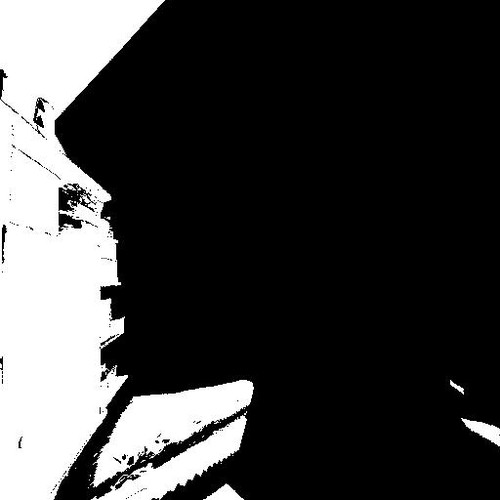}
    \includegraphics[width=0.235\linewidth]{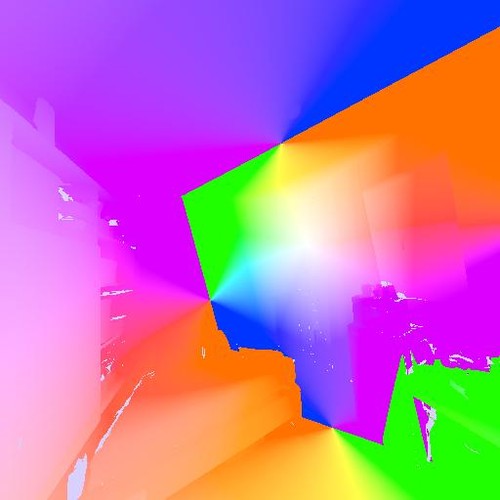}

    \includegraphics[width=0.235\linewidth]{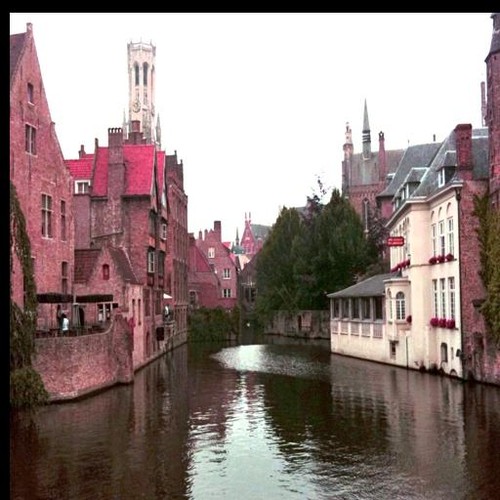}
    \includegraphics[width=0.235\linewidth]{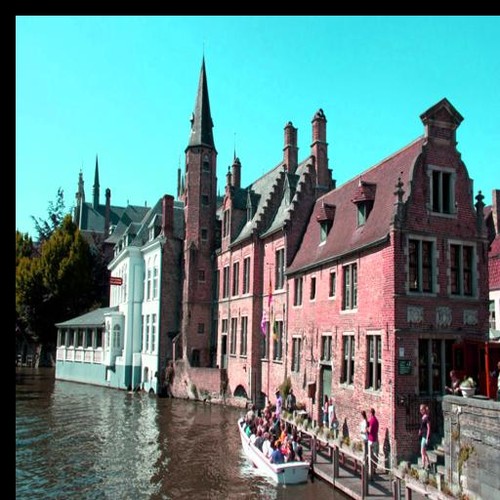}
    \includegraphics[width=0.235\linewidth]{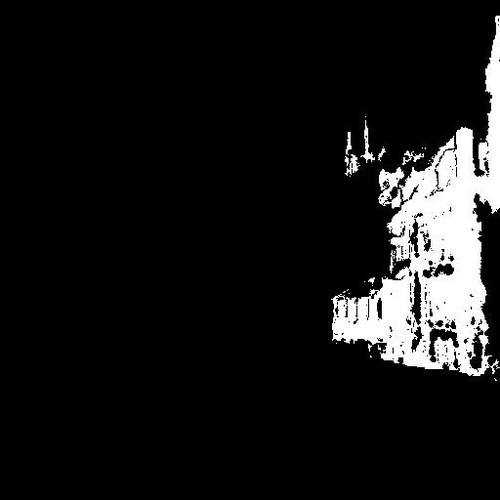}
    \includegraphics[width=0.235\linewidth]{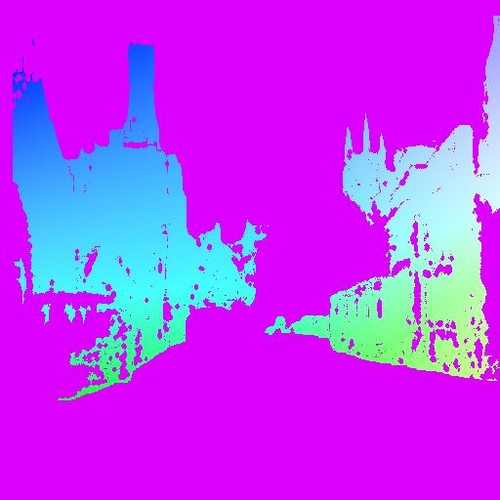}

    \includegraphics[width=0.235\linewidth]{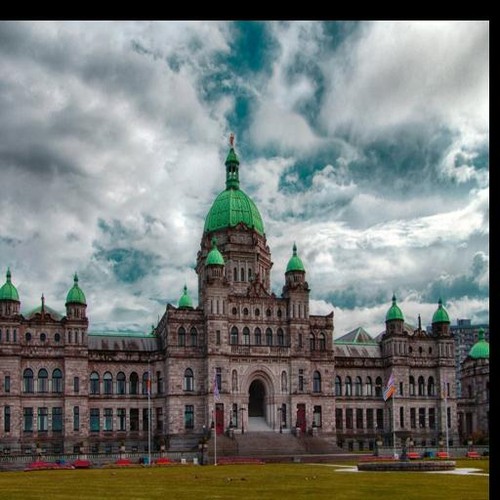}
    \includegraphics[width=0.235\linewidth]{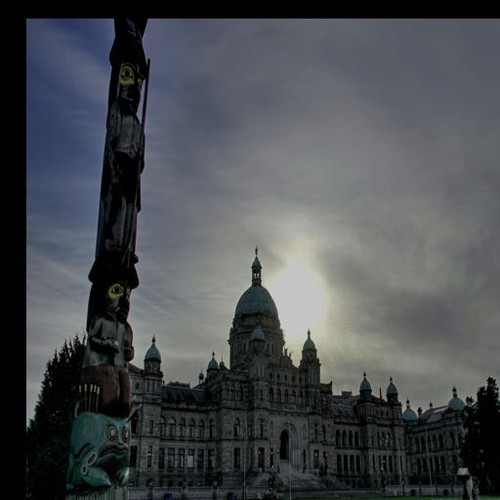}
    \includegraphics[width=0.235\linewidth]{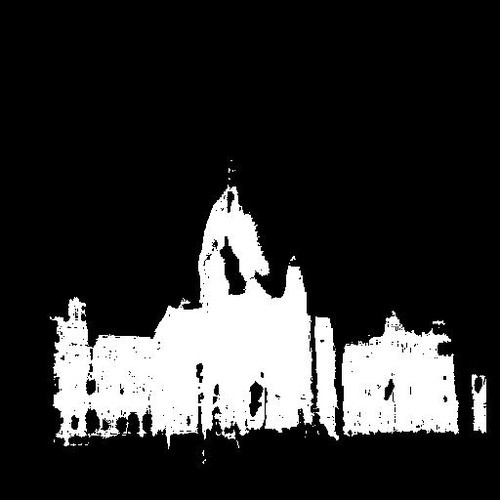}
    \includegraphics[width=0.235\linewidth]{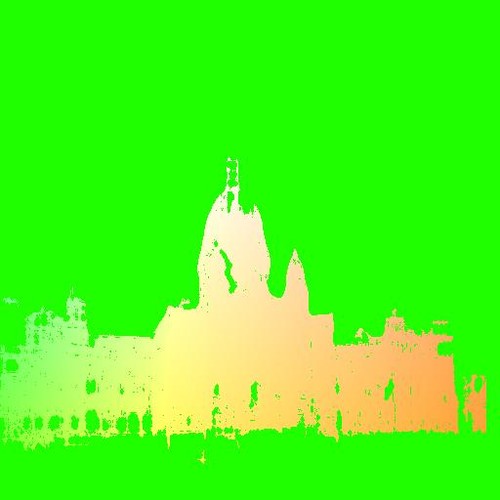}

    \includegraphics[width=0.235\linewidth]{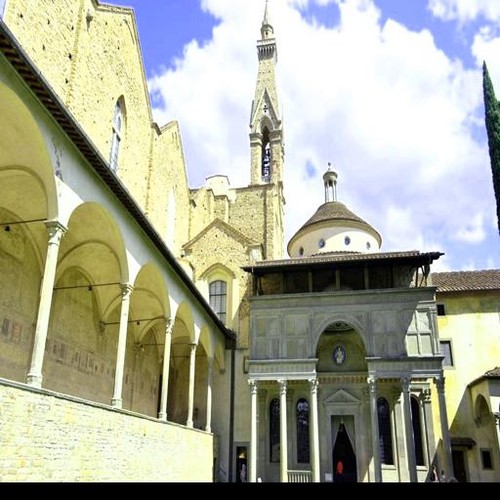}
    \includegraphics[width=0.235\linewidth]{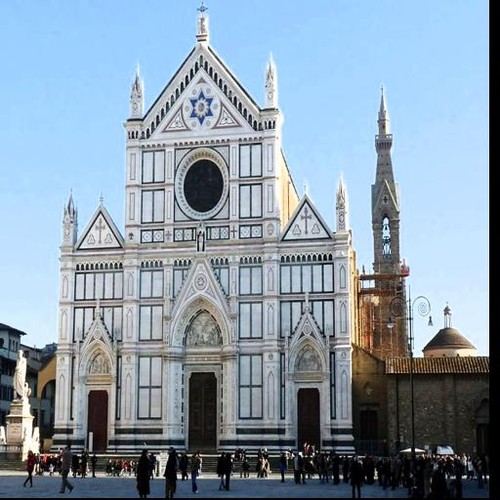}
    \includegraphics[width=0.235\linewidth]{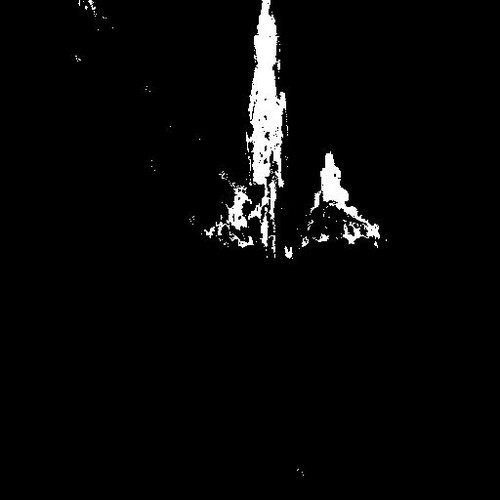}
    \includegraphics[width=0.235\linewidth]{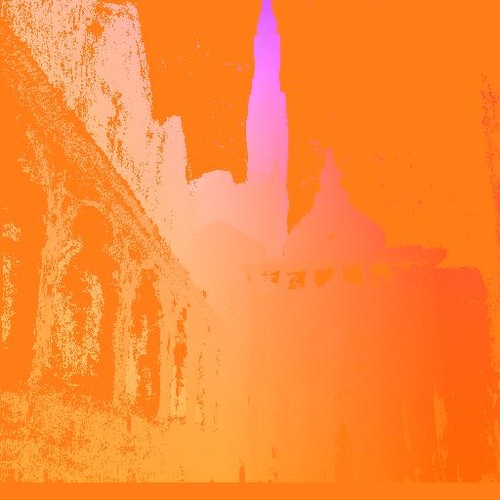}

    \includegraphics[width=0.235\linewidth]{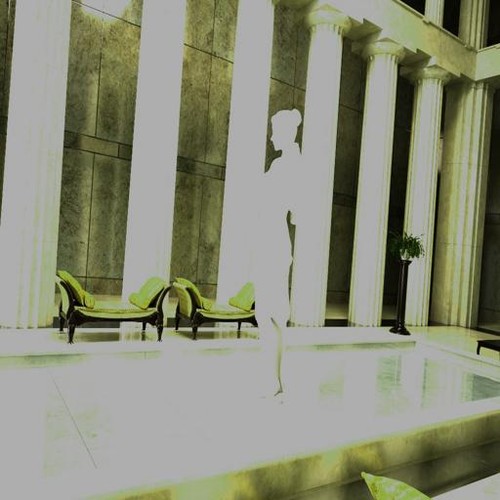}
    \includegraphics[width=0.235\linewidth]{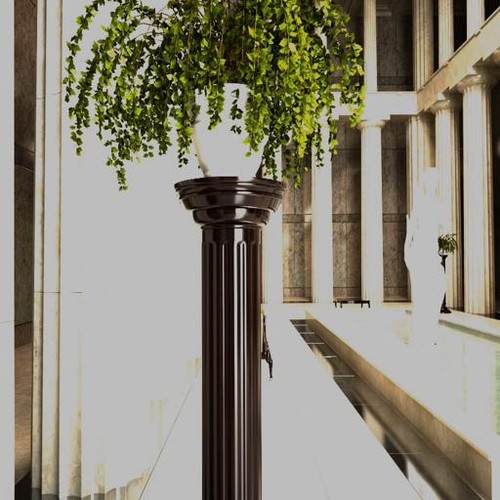}
    \includegraphics[width=0.235\linewidth]{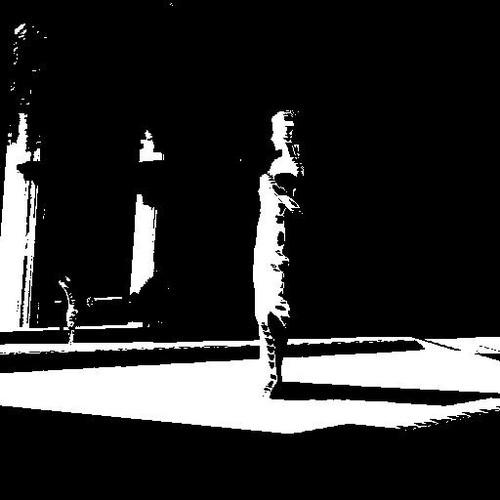}
    \includegraphics[width=0.235\linewidth]{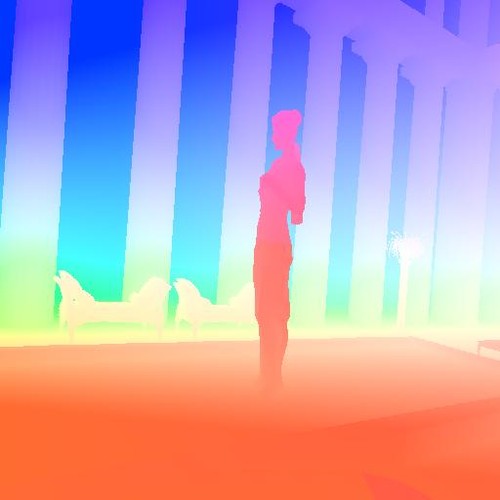}

    \includegraphics[width=0.235\linewidth]{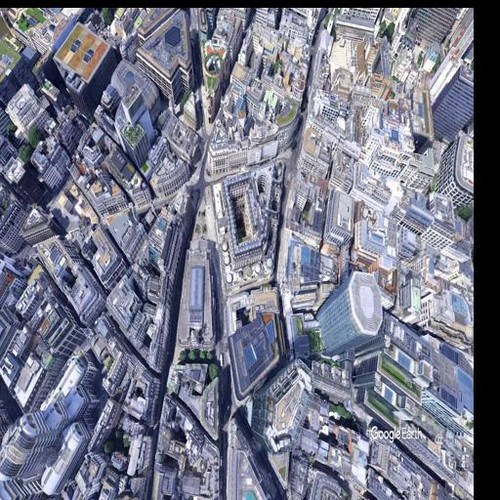}
    \includegraphics[width=0.235\linewidth]{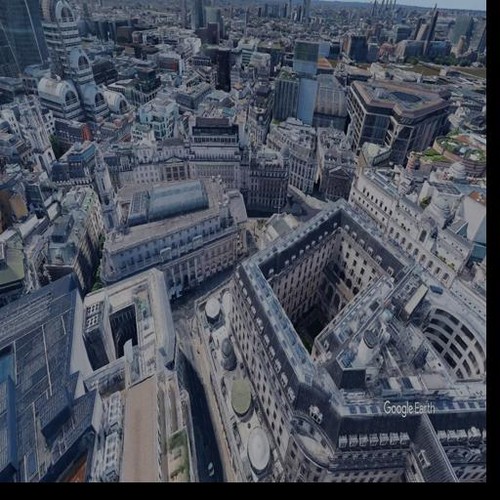}
    \includegraphics[width=0.235\linewidth]{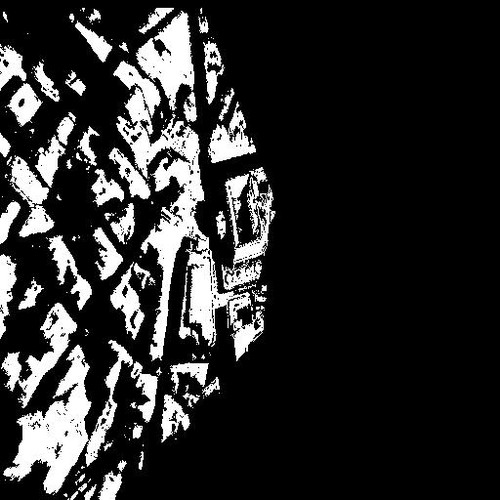}
    \includegraphics[width=0.235\linewidth]{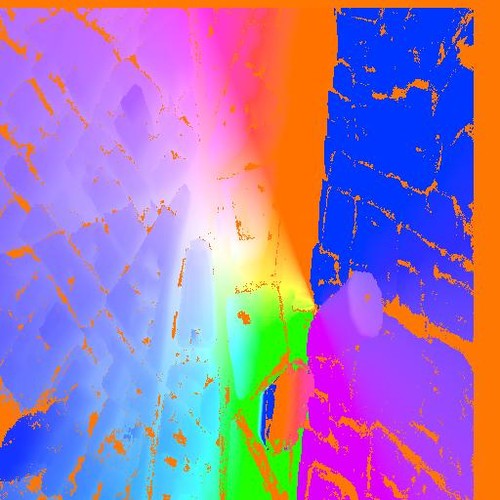}

    \includegraphics[width=0.235\linewidth]{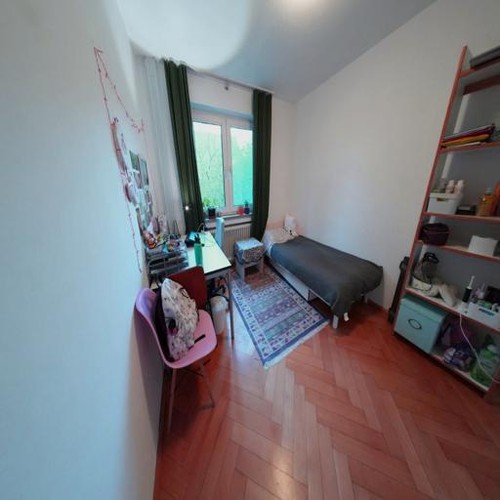}
    \includegraphics[width=0.235\linewidth]{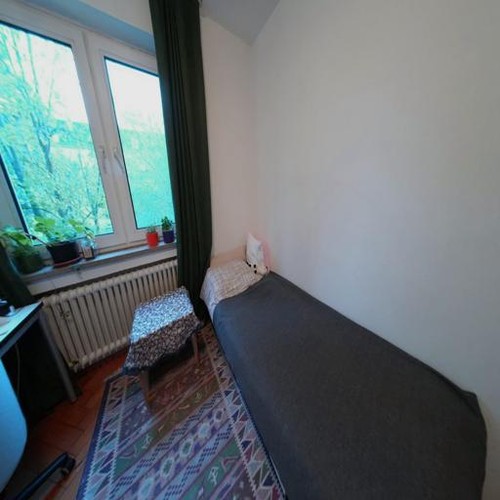}
    \includegraphics[width=0.235\linewidth]{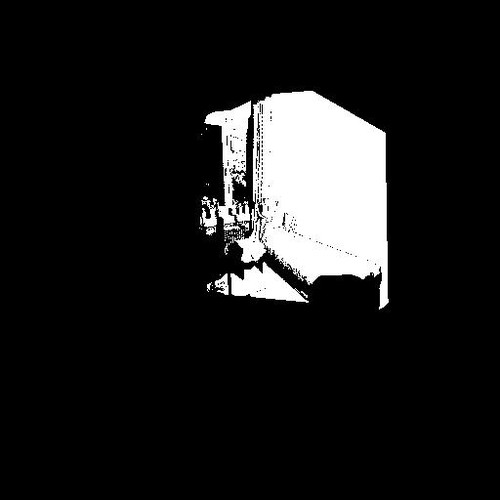}
    \includegraphics[width=0.235\linewidth]{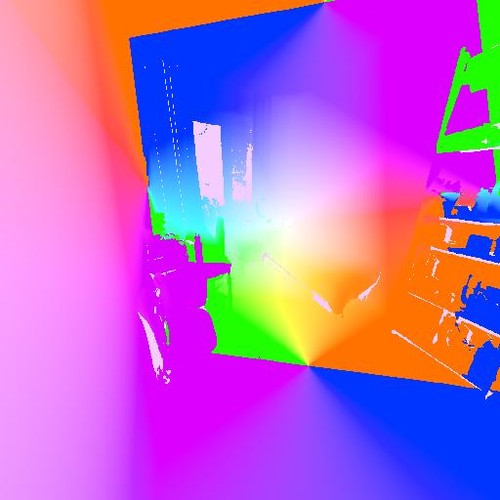}

\end{subfigure}%
\hfill 
\begin{subfigure}[t]{0.49\linewidth}
    \centering
    \includegraphics[width=0.235\linewidth]{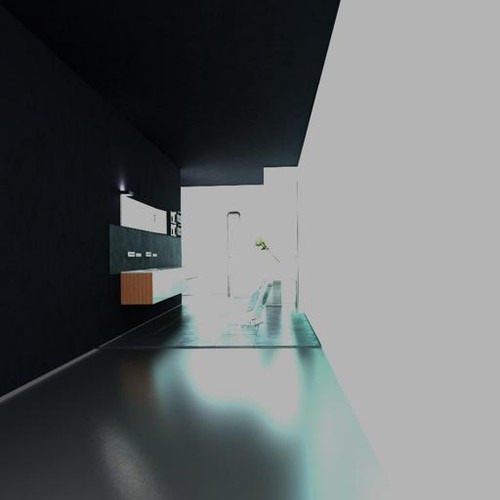}
    \includegraphics[width=0.235\linewidth]{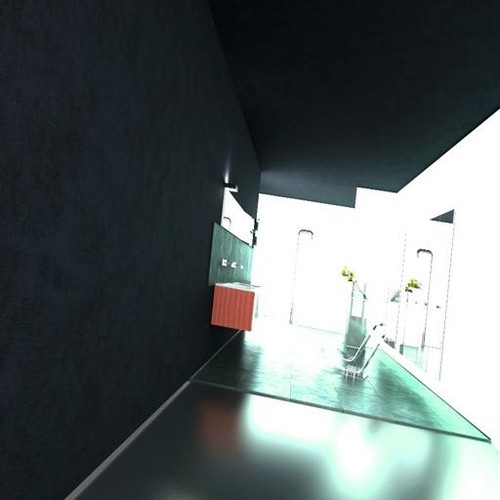}
    \includegraphics[width=0.235\linewidth]{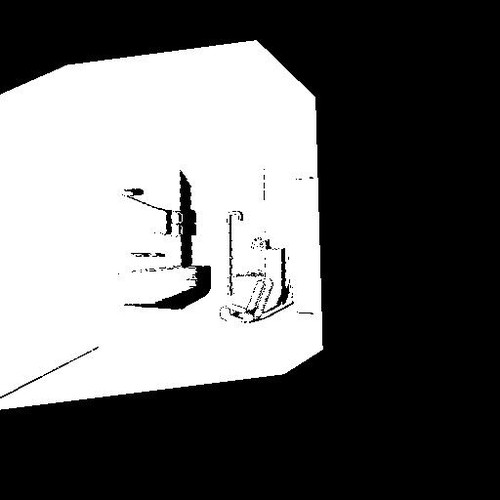}
    \includegraphics[width=0.235\linewidth]{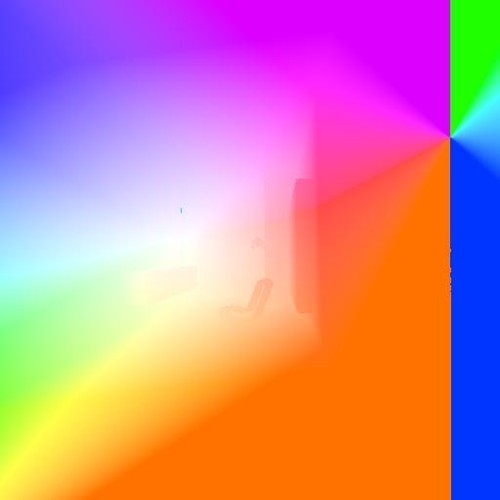}

    \includegraphics[width=0.235\linewidth]{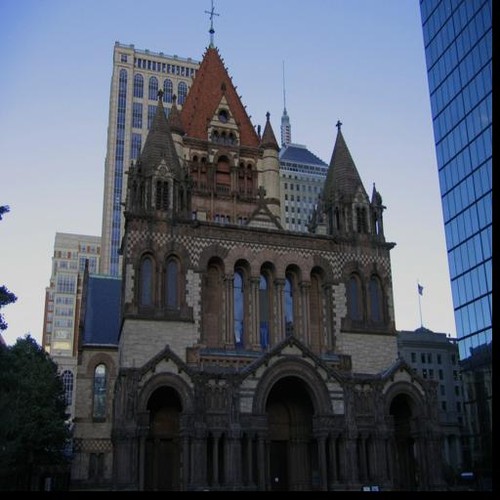}
    \includegraphics[width=0.235\linewidth]{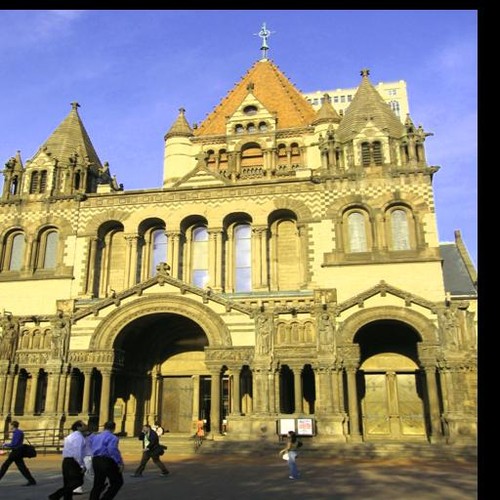}
    \includegraphics[width=0.235\linewidth]{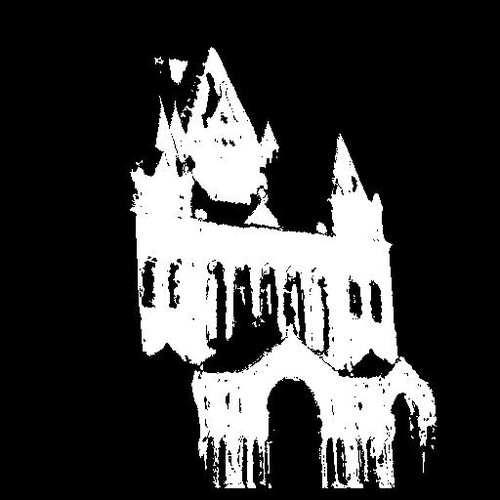}
    \includegraphics[width=0.235\linewidth]{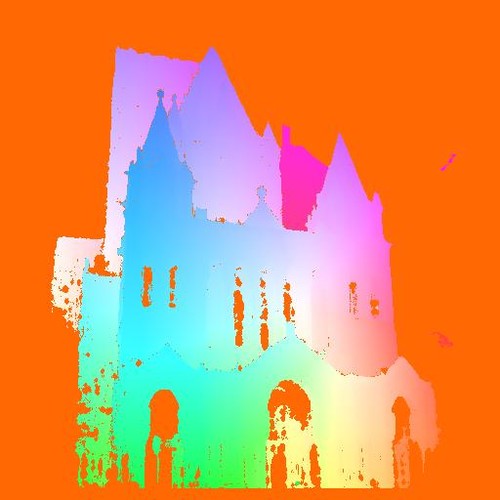}

    \includegraphics[width=0.235\linewidth]{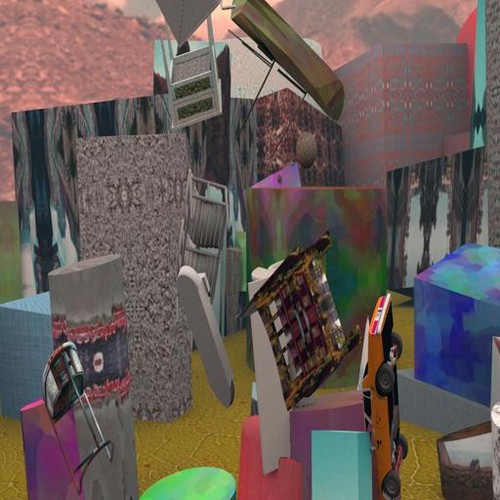}
    \includegraphics[width=0.235\linewidth]{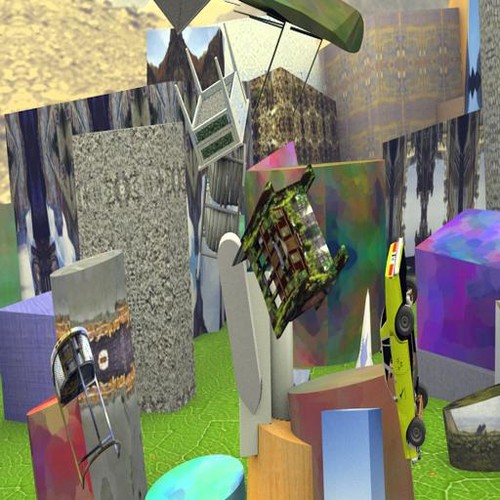}
    \includegraphics[width=0.235\linewidth]{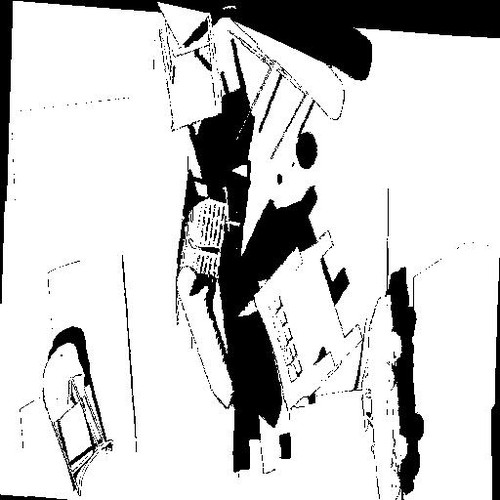}
    \includegraphics[width=0.235\linewidth]{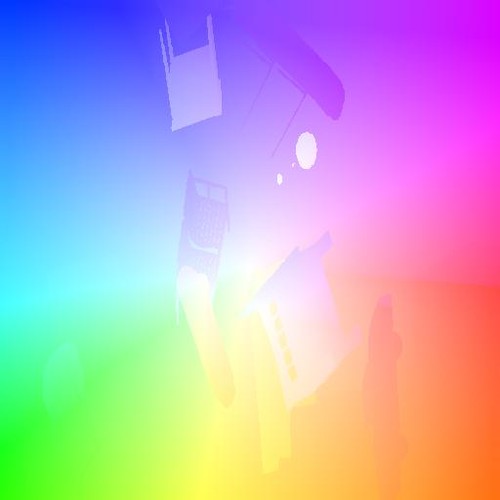}

    \includegraphics[width=0.235\linewidth]{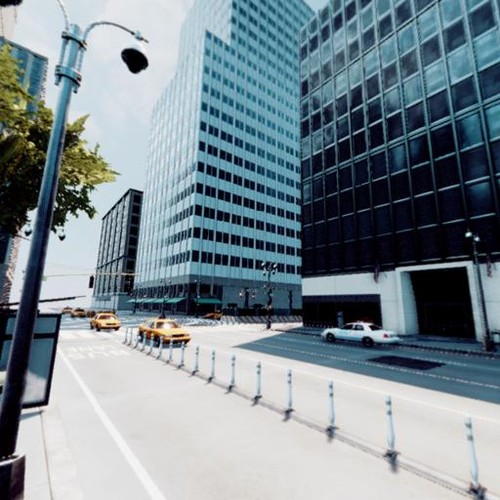}
    \includegraphics[width=0.235\linewidth]{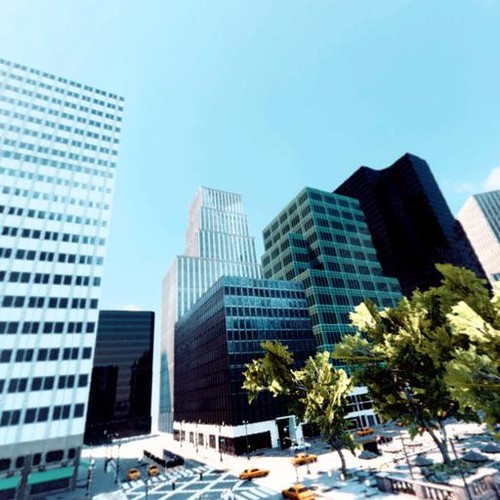}
    \includegraphics[width=0.235\linewidth]{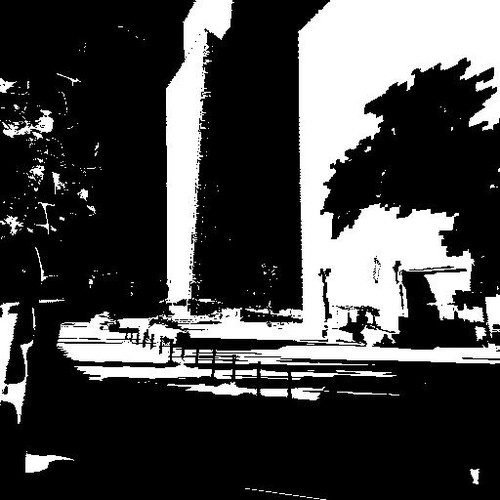}
    \includegraphics[width=0.235\linewidth]{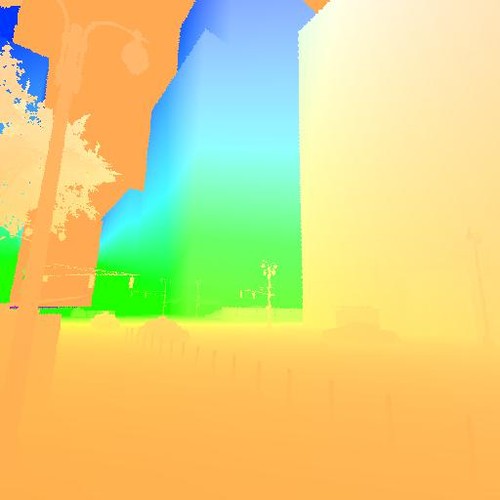}

    \includegraphics[width=0.235\linewidth]{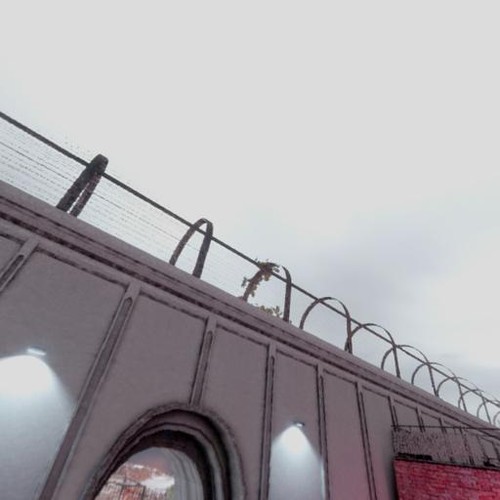}
    \includegraphics[width=0.235\linewidth]{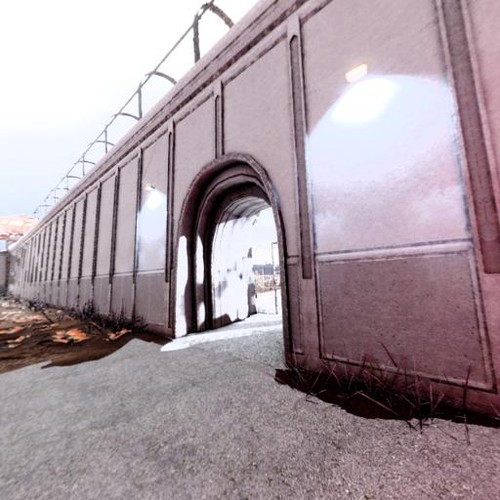}
    \includegraphics[width=0.235\linewidth]{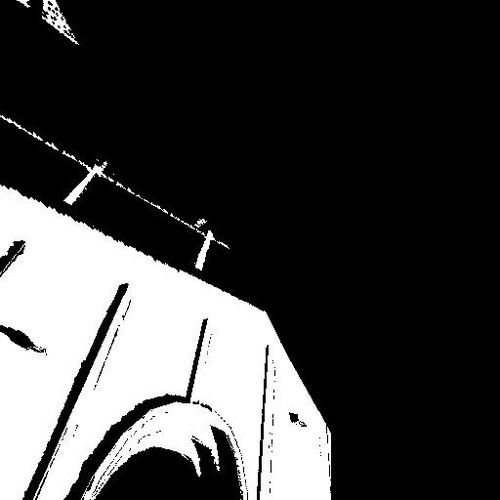}
    \includegraphics[width=0.235\linewidth]{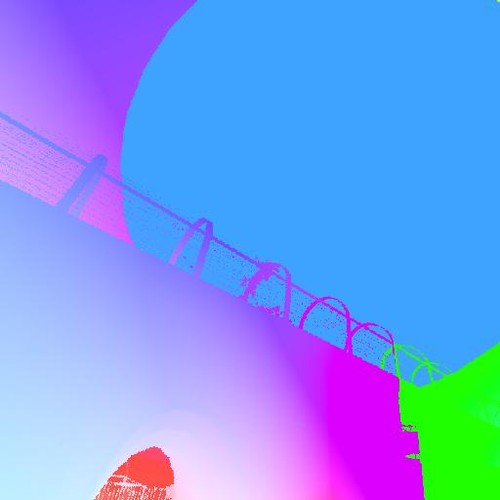}

    \includegraphics[width=0.235\linewidth]{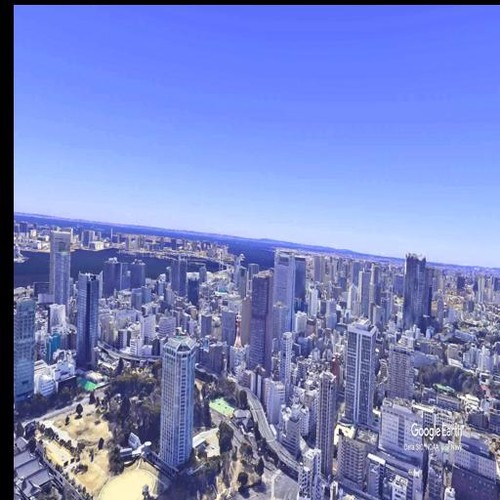}
    \includegraphics[width=0.235\linewidth]{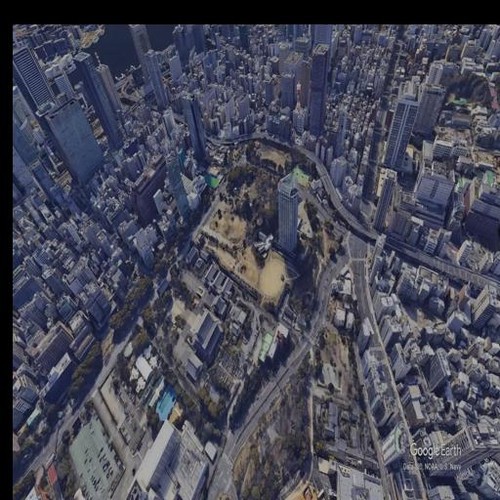}
    \includegraphics[width=0.235\linewidth]{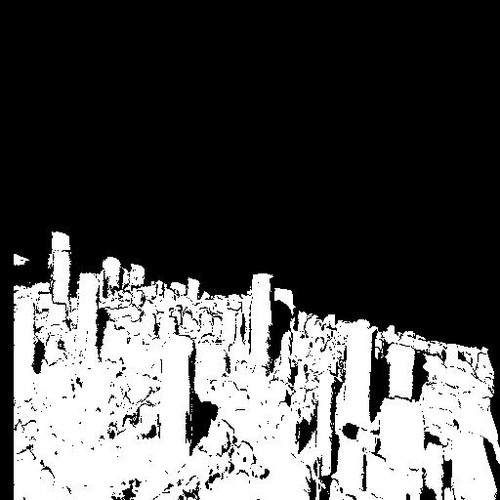}
    \includegraphics[width=0.235\linewidth]{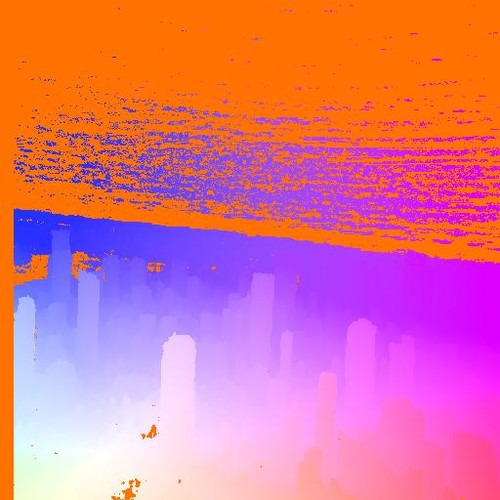}

    \includegraphics[width=0.235\linewidth]{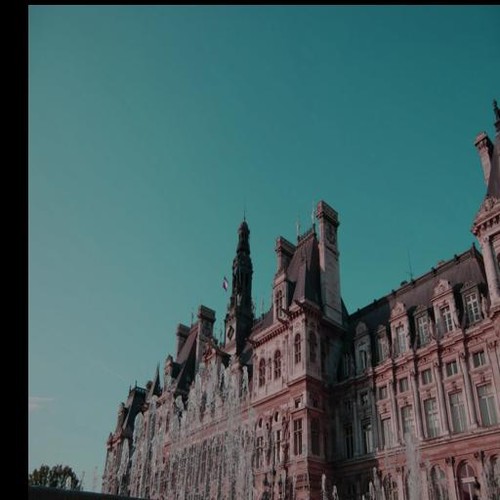}
    \includegraphics[width=0.235\linewidth]{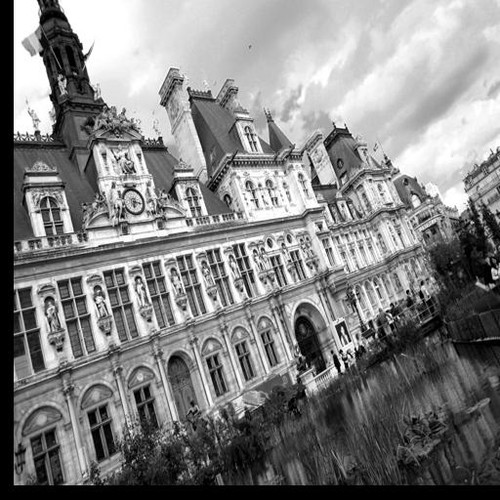}
    \includegraphics[width=0.235\linewidth]{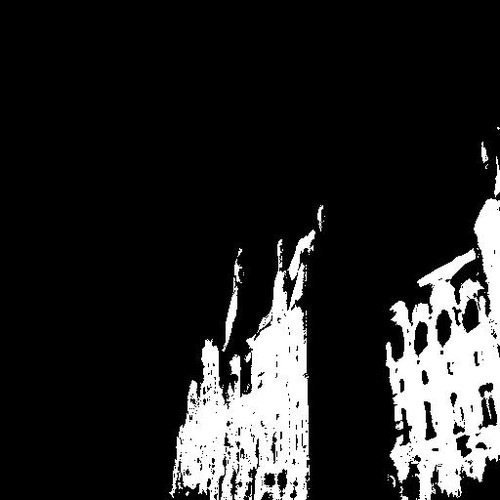}
    \includegraphics[width=0.235\linewidth]{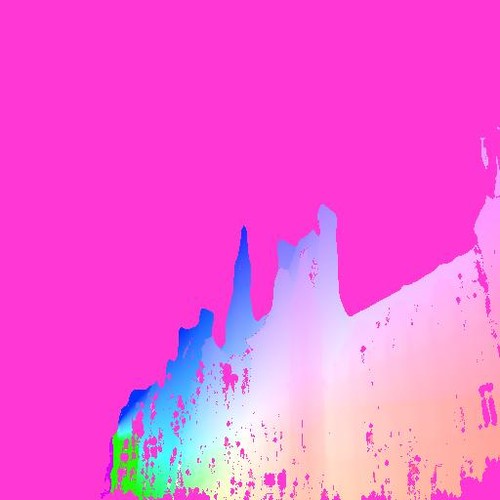}

    \includegraphics[width=0.235\linewidth]{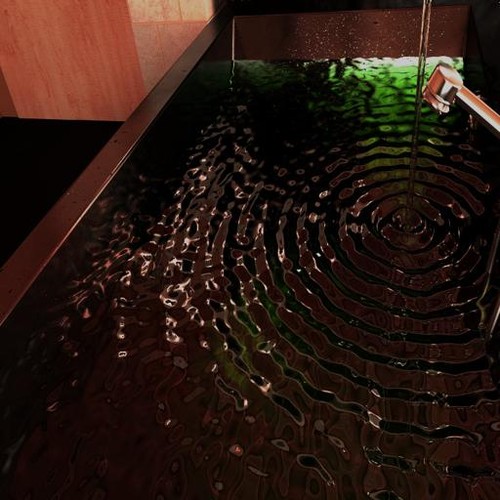}
    \includegraphics[width=0.235\linewidth]{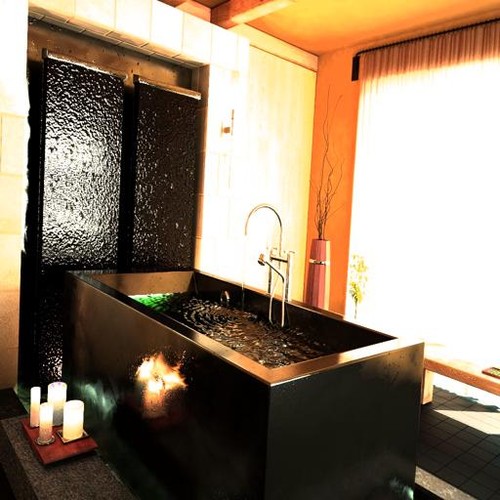}
    \includegraphics[width=0.235\linewidth]{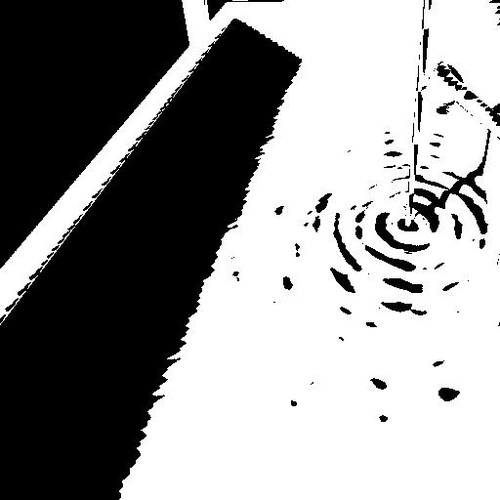}
    \includegraphics[width=0.235\linewidth]{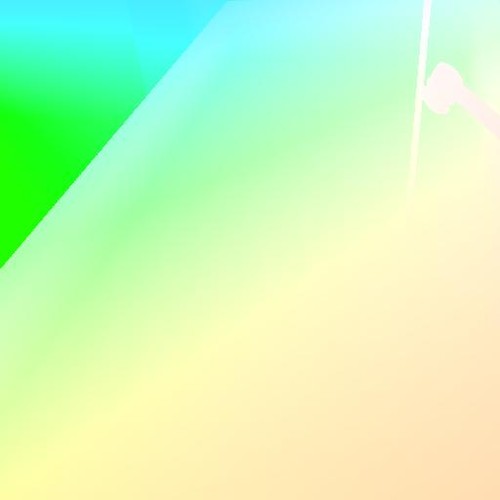}

\end{subfigure}
\caption{\textbf{Visualization of training batch.} Our data mixure is diverse and challenging, covering many types of scenes.}
\label{fig:train-batch}
\end{figure*}

\paragraph{Augmentations:}
Besides using different aspect ratios, we additionally employ light data augmentation.
Specifically, we use horizontal flipping, grayscale with probability 0.1, multiplicative brightness (ratio between [1/1.5, 1.5]), and hue jitter ($[-15^{\circ}, 15^{\circ}]$ in the HSV parameterization).
For MegaDepth and AerialMegaDepth we additionally follow RoMa and translate the image randomly in the range $\{-32, ..., 32\}$ in both rows and columns.
\paragraph{Visualization of Training Batch:}
We visualize a randomly drawn batch in from the training data in~\Cref{fig:train-batch}, in order to give a qualitative understanding of the type of pairs \ours~is trained on.

\paragraph{Overlap/Covisibility Computation:}
For depth-based datasets (all datasets except FlyingThings3D, UnrealStereo4k, and Virtual KITTI 2) we use depth consistency to compute pixel-wise covisibility.
We say that the depth is consistent if
\begin{equation}
    \frac{|\mathbf{z}^{B}(x^{A\mapsto B}) - z^{A\mapsto B}|}{ \mathbf{z}^{B}(x^{A\mapsto B})} < \tau = 0.05,
\end{equation}
where $x^{A\mapsto B}$ is the mapping of the pixel-coordinate $x^A$ into $B$ as $x^{A\mapsto B} \sim K^B(R^{A\mapsto B}(K^A)^{-1}x^A + t)$ and $z^{A\mapsto B} = (K^B(R^{A\mapsto B}(K^A)^{-1}x^A + t))_3$ is the corresponding depth.

For flow-based datasets, we measure the warp cycle consistency as 
\begin{equation}
    \norm[\big]{\mathbf{W}^{B\mapsto A}(\mathbf{W}^{A\mapsto B}(x^A)) - x^A} < 5\cdot 10^{-3} \approx 1.6 \text{ px}
\end{equation}
at a resolution of $640\times 640$.
\section{Further Details on Benchmarks}
\label{app:benchmark_details}
\paragraph{SatAst:}
We create a new matching benchmark called SatAst (Satellite, Astronaut), that uses images taken by astronauts from the international space station and satellite images.
We take 39 pairs of corresponding images from EarthMatch~\cite{berton2024earthmatch} (which they in turn took from AIMS~\cite{stoken2023astronaut}).
The pairs in EarthMatch were obtained by retrieving ten satellite images from a large database for each given astronaut image.
For our benchmark we only select image pairs that are correctly matching (confirmed by visual inspection).
We also exclude images with extreme cloud occlusions as well as images where we were not able to accurately annotate correspondences that agree on a homography.

\begin{figure*}
    \centering
    \includegraphics[width=0.75\linewidth]{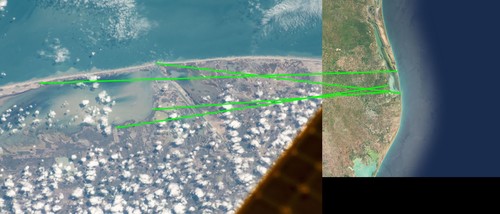}
    \includegraphics[width=0.75\linewidth]{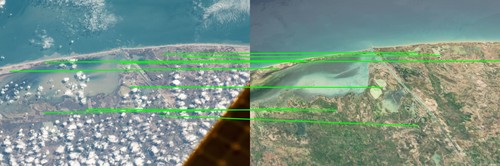}
    \includegraphics[width=0.75\linewidth]{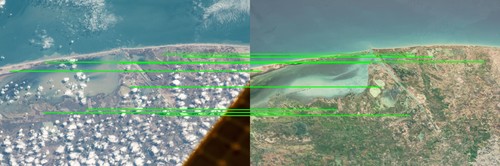}
    \includegraphics[width=0.75\linewidth]{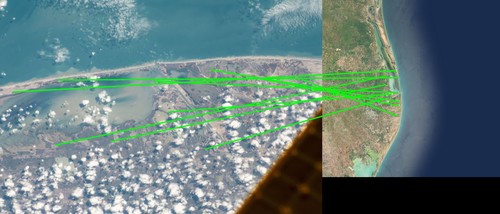}
    \caption{\textbf{Annotation of correspondences for SatAst:} 1) We annotate four initial approximate correspondences. 2) We warp the satellite image using the homography obtained from the previous step and annotate ten accurate correspondences. 3) Visualization of the warp obtained by estimating a homography from the ten accurate annotations. 4) The ten accurate correspondences visualized in the original images, where we score the homographies obtained from the dense matchers. Step 2) is sometimes repeated several times until a warp that is deemed good enough is obtained.}
    \label{fig:satast}
\end{figure*}

We annotate corresponding points in the image pairs in an iterative fashion as illustrated in Figure~\ref{fig:satast}.

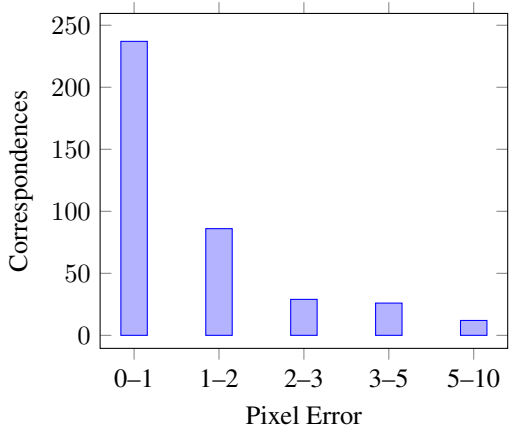
\begin{figure}
    \centering
    \begin{tikzpicture}
        \begin{axis}[
            ybar,
            ylabel={Correspondences},
            ytick={0,50,100,150,200,250},
            xlabel={Pixel Error},
            symbolic x coords={0--1,1--2,2--3,3--5,5--10},
            xtick=data,
            ]
        \addplot coordinates {
            (0--1,237) 
            (1--2,86) 
            (2--3,29)
            (3--5,26)
            (5--10,12) 
        };
        \end{axis}
    \end{tikzpicture}
    \caption{\textbf{Accuracy of annotations on SatAst:} A histogram over the reprojection errors of the $390$ annotated correspondences in SatAst according to homographies estimated from the ten annotations in each image. The image resolution of the satellite images is $3072\times 3072$, so an error of $10$ pixels is around $0.3\%$ of the image width.}
    \label{fig:satast-hist}
\end{figure}

To get a sense of how good the annotations are, we estimate homographies from them and calculate the reprojection error from mapping the points from the astronaut image through the homography to the satellite image.
The resulting errors are plotted in Figure~\ref{fig:satast-hist}.

\section{Further Details on Predictive Covariance Experiment}
We create a benchmark out of 1500 pairs from validation scenes the HyperSim~\cite{roberts2021hypersim} dataset, where pairs with $<0.2$ overlap are discarded.

Since \ours~predicts only the forward covariance (and the residuals are two-sided), we approximate the full $4\times 4$ covariance matrix of the matches by a block diagonal matrix where for each drawn correspondence pair $x^A, x^B$ the covariance of points in $\mathbf{I}^A$ are approximated by sampling the backwards covariance as
\begin{equation}
    (\mathbf{\Sigma}^{-1})^{A}(x^A) \approx (\mathbf{\Sigma}^{-1})^{B\mapsto A}(\mathbf{W}^{A\mapsto B}(x^A)).
\end{equation}

We optimize the covariance weighted Sampson error~\cite{chojnacki2000fitting},
\begin{equation}
    \frac{ (\boldsymbol{x}_B^T F \boldsymbol{x}_A) }{ \| F_{12} \boldsymbol{x}_A\|^2_{\Sigma_A} + \| (F^T)_{12} \boldsymbol{x}_B\|^2_{\Sigma_B} }
\end{equation}
where $\|\boldsymbol{u}\|_\Sigma^2 = \boldsymbol{u}^T \Sigma \boldsymbol{u}$. 
For the robust estimation experiment, we similarly use the residual inside the MSAC scoring, inside a standard LO-RANSAC.

\section{Relative Pose Estimation}
\label{suppl:relpose}

\paragraph{VGGT:}
For evaluating VGGT on MegaDepth-1500, we follow the evaluation outlined by VGGT~\cite{wang2025vggt}. In particular, we sample 1024 keypoints using ALIKED~\cite{Zhao2023ALIKED} and use as query points in the tracking head. We try different confidence and covisibility thresholds. We settle for $0.1$ for both. 

\paragraph{UFM:}
We follow the same sampling as for~\ours~and RoMa, using bidirectional warps and balanced sampling.
However, as UFM only supports a fixed resolution of $H=420 \times W=560$ we do not use upsampling.

\section{Bias In AerialMegaDepth}
\begin{figure}
    \centering
    \includegraphics[width=0.49\linewidth]{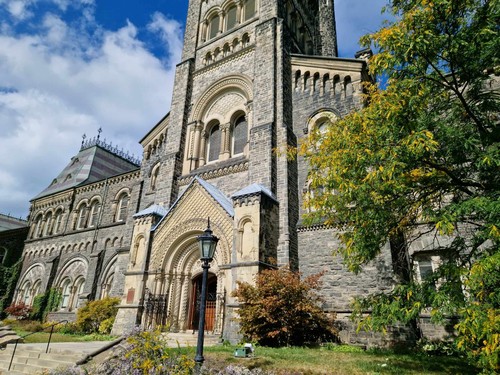}
    \includegraphics[width=0.49\linewidth]{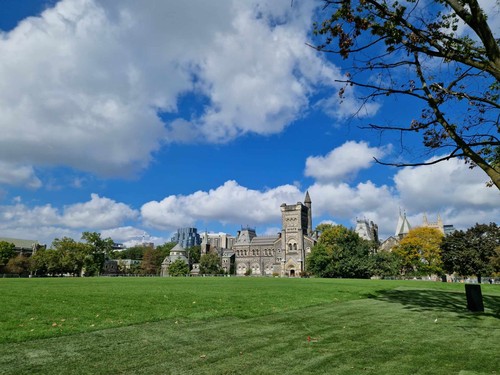}
    \includegraphics[width=\linewidth]{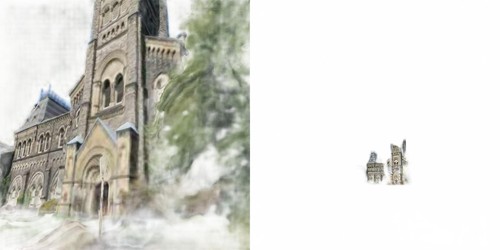}    
    \caption{\textbf{Visualization of \ours~warp.} Note that the model puts some confidence erroneously in sky pixels (see top-left). This may be due to bias stemming from AerialMegaDepth.}
    \label{fig:sky-bias}
\end{figure}

\begin{figure}
    \centering
    \includegraphics[width=0.49\linewidth]{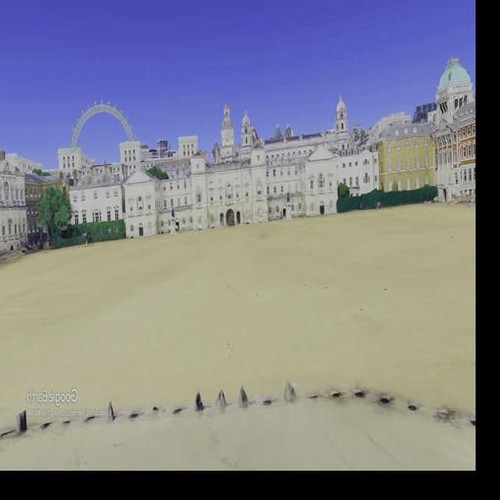}
    \includegraphics[width=0.49\linewidth]{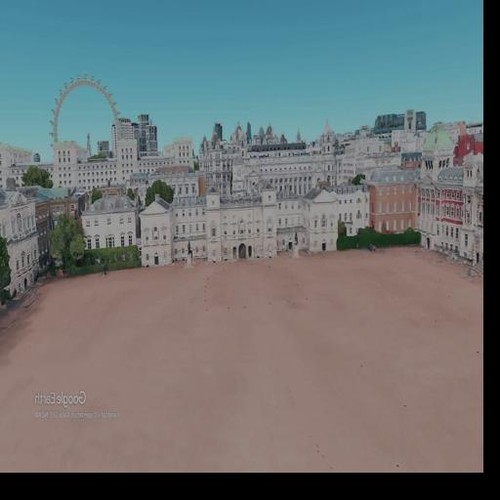}
    \includegraphics[width=0.49\linewidth]{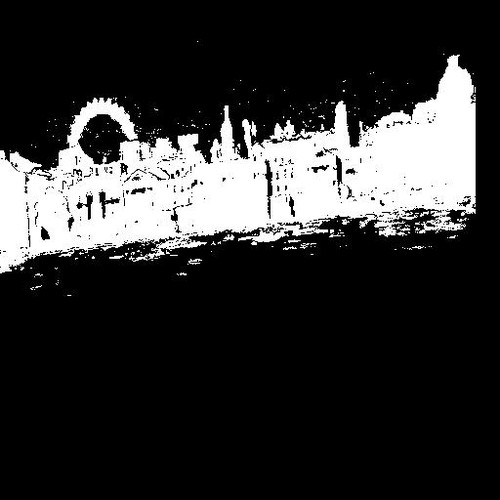}
    \includegraphics[width=0.49\linewidth]{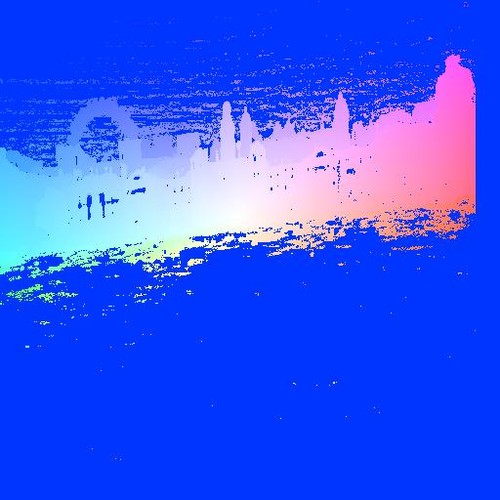}
    \caption{\textbf{Spurious depth estimates in AerialMegaDepth.} Depth from the scene leaks into the sky, causing some sky pixels to be multi-view consistent. This possibly leaks into the warp estimate of \ours.}
    \label{fig:aerialmd-bias}
\end{figure}

We observe that \ours~sometimes predicts overlaps in textureless sky regions, as demonstrated in~\Cref{fig:sky-bias}
 We believe that this is due to AerialMegaDepth sometimes propagating depths from the mesh to sky pixels, as illustrated in~\Cref{fig:aerialmd-bias}.

\bibliographystyle{splncs04}
\end{document}